\documentclass[lettersize,journal]{IEEEtran}
\usepackage{amsmath,amsfonts}
\usepackage{algorithmic}
\usepackage{algorithm}
\usepackage{array}

\ifCLASSOPTIONcompsoc
\usepackage[caption=false,font=normalsize,labelfont=sf,textfont=sf]{subfig}
\else
\usepackage[caption=false,font=footnotesize]{subfig}
\fi

\makeatletter
\let\NAT@parse\undefined
\makeatother
\usepackage{hyperref}  %hyperref still needs to be put at the end!
\hypersetup{hidelinks}

\usepackage{textcomp}
\usepackage{stfloats}
\usepackage{url}
\usepackage{verbatim}
\usepackage{graphicx}
\usepackage{float} 
\usepackage{color}
\usepackage{cite}
\usepackage{bm}
\usepackage{booktabs}
\usepackage{multirow}
\usepackage{tabularx}
\graphicspath{{pictures/}}
\allowdisplaybreaks[4]
\begin{document}

\title{\footnotesize This work has been submitted to the IEEE for possible publication.\\Copyright may be transferred without notice, after which this version may no longer be accessible.\\
\Huge
Human-aligned Safe Reinforcement Learning for Highway On-Ramp Merging in Dense Traffic}

\author{Yang Li, Shijie Yuan, Yuan Chang, Xiaolong Chen, Qisong Yang, Zhiyuan Yang, Hongmao Qin

\thanks{This work was supported by NSF of China with 52302493, 52102394, and 52172384, Hunan Provincial NSF of China with 2021JJ40095, and State Key Laboratory of Advanced Design and Manufacturing for Vehicle Body with 61775006. (\textit{Corresponding author: Hongmao Qin.})}
\thanks{Yang Li, Shijie Yuan, Yuan Chang, Xiaolong Chen, Zhiyuan Yang, and Hongmao Qin are with the College of Mechanical and Vehicle Engineering, Hunan University, Changsha 410082, China. (e-mail: lyxc56@gmail.com; yuansj@hnu.edu.cn; cy00@hnu.edu.cn; xlchan18@hnu.edu.cn; yangzhy@hnu.edu.cn; qinhongmao@hnu.edu.cn).}
\thanks{Qisong Yang is with the Algorithmics Group of the EEMCS, Delft University of Technology, The Netherlands. (e-mail: q.yang@tudelft.nl).}}% <-this % stops a space

% The paper headers
%\markboth{IEEE Intelligent Transportation System Magazine}%
\markboth{}%
{Shell \MakeLowercase{\textit{et al.}}: A Sample Article Using IEEEtran.cls for IEEE Journals}

\maketitle

%% abstract
\begin{abstract}
Most reinforcement learning (RL) approaches for the decision-making of autonomous driving consider safety as a reward instead of a cost, which makes it hard to balance the tradeoff between safety and other objectives. Human risk preference has also rarely been incorporated, and the trained policy might be either conservative or aggressive for users. To this end, this study proposes a human-aligned safe RL approach for autonomous merging, in which the high-level decision problem is formulated as a constrained Markov decision process (CMDP) that incorporates users' risk preference into the safety constraints, followed by a model predictive control (MPC)-based low-level control. The safety level of RL policy can be adjusted by computing cost limits of CMDP's constraints based on risk preferences and traffic density using a fuzzy control method. To filter out unsafe or invalid actions, we design an action shielding mechanism that pre-executes RL actions using an MPC method and performs collision checks with surrounding agents. We also provide theoretical proof to validate the effectiveness of the shielding mechanism in enhancing RL's safety and sample efficiency. Simulation experiments in multiple levels of traffic densities show that our method can significantly reduce safety violations without sacrificing traffic efficiency. Furthermore, due to the use of risk preference-aware constraints in CMDP and action shielding, we can not only adjust the safety level of the final policy but also reduce safety violations during the training stage, proving a promising solution for online learning in real-world environments. 
\end{abstract}

\section{Introduction}
\begin{figure}[!t]
\centering
\includegraphics{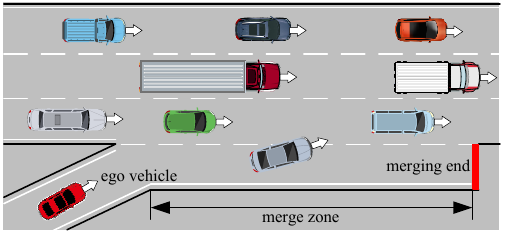}
\caption{Illustration of the highway on-ramp merging scenario. The ego vehicle (green) starts on a one-lane entrance ramp and needs to merge into the highway traffic safely and efficiently.}
\label{fig:merge scenario}
\end{figure}

%(安全和效率tradeoff, 运动预测，风险可调)
\IEEEPARstart{M}{erging} into the congested highway traffic has been challenging for autonomous vehicles (AVs) since the decision-making process must handle two conflicting goals, \textit{i.e.}, safety and efficiency \cite{Nishi2019, chen2023, huang2023general}. An over-conservative decision is safe enough but may cause the ``freezing robot" problem \cite{Trautman2010}, and an aggressive decision gains efficiency but sacrifices safety. Reaching a harmonious balance between safety and efficiency is essential, especially in dense traffic situations. Besides, human drivers may make different decisions even in the same situations, for instance, the conservative driver usually chooses risk-averse decisions and it may take a longer time to merge, while the aggressive driver often chooses actions that can make the merge faster but a little bit risky. To accommodate the personal preferences of different users, it is crucial to consider individual differences regarding risk perception in the decision-making process of AVs. In recent years, reinforcement learning (RL) has attracted increasing attention in AVs \cite{Nishi2019, Wang2017, Lin2020, Wang2021, Aradi2022, kamran2022modern, aboyeji2023effect, he2023fear, du2023safe,hou2023subtask, hou2023exploring, deniz2024reinforcement,zhang2024analyzing}, which trains an agent to learn the optimal decisions by trial and error \cite{degrave2022magnetic, orlowski2023multiagent, crewe2023slav}. Safety is a critical concern, and most studies formulate safety specifications into reward functions, which cannot yet guarantee safety in the learned policy \cite{Lubars2021}. Also, safety violations are unavoidable in the exploration process. Some works use adversarial techniques to enhance the robustness of RL methods\cite{he2023fear, he2023towards,yang2023cem}. Safe exploration aims to train the agent to avoid dangerous actions during the training process \cite{koller2018learning}, and it has been rarely studied. Currently, the sim-to-real gap and difficulties in building a high-fidelity simulator are likely to boost the need for real-world training, and actions that can cause damage would not be allowed during the training process. In addition, most works do not consider personal preferences on safety, and thus, the learned RL policy might not align well with the user's risk perception, leading to reduced satisfaction and mistrust \cite{lyu2021}. Therefore, this study aims to propose a human-aligned safe RL framework for the decision-making of the highway on-ramp merging task of AVs, as shown in Fig. \ref{fig:merge scenario}. The ego vehicle (green) starts on a one-lane entrance ramp and needs to merge into the highway traffic safely and efficiently.
%% Illustration of the highway on-ramp merging scenario
%% Overview of the proposed method
\begin{figure*}[!t]
\centering
\includegraphics[width=\linewidth]{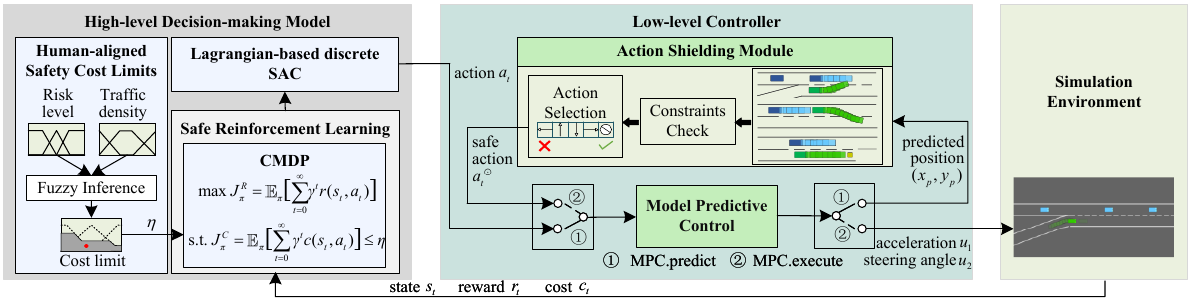}
\caption{{Overview of the proposed method. 
The high-level decision problem is formulated as a CMDP that incorporates individuals' risk preferences into the constraints, followed by an MPC-based low-level control. A Lagrangian-based SAC algorithm is used to solve CMDP for the optimal RL policy. The RL action is a discrete action such as left, right, acceleration, etc. To filter out unsafe or invalid RL actions, we design an action shielding mechanism to mask out risky ones by pre-executing the action with MPC and conducting collision constraint checks. Then, the safe RL action is sent to the low-level MPC, which generates the vehicle control (the acceleration and steering angle) for the simulation environment. The simulation environment finally generates the state, reward, and cost data for training the high-level RL agent. The RL agent learns to act by trial and error, and the safety violations during the training process can be reduced using our method. }}
\label{srl mpc architecture}
\end{figure*}
%%%%% Motivations and Contributions
\subsection{Motivations and Contributions}
RL has gained wide attention for the on-ramp merging task of autonomous driving, and it should balance safety and efficiency to enable the car to successfully merge into the dense traffic. Many studies consider safety in the reward function instead of constrained setting, and safety violations still exist during the exploration and execution phases. Besides, individual differences in risk perception are rarely studied, and the learned RL policy usually fails to match the needs of different users. {Therefore, this study proposes a human-aligned safe RL approach for the on-ramp merging of autonomous driving, as shown in Fig. \ref{srl mpc architecture}. We formulate the high-level decision-making problem as a constrained Markov decision process (CMDP), where safety specifications are formulated as constraints that consider human risk preference. We solve the high-level safe RL problem using a Lagrangian-based soft actor-critic (SAC) algorithm to find the optimal RL policy. To remove unsafe or invalid RL actions, an action shielding module (ASM) is built, and we pre-execute RL actions in the Model Predictive Control (MPC) module and then conduct collision checks with surrounding agents to mask out risky ones. The selected safe RL action is finally sent to MPC for low-level trajectory planning. Moreover, we conduct theoretical analysis and numerical simulations to validate the proposed method's safety performance and convergence rate.}

{Different from \cite{Lubars2021,albarella2023hybrid,chen2023}, we focus on single-agent RL and use a CMDP model that incorporates human risk preferences to adapt to the users' specific expectations of safety. The safety level of the RL policy can accommodate different users by determining the cost limits of CMDP using a fuzzy decision method. Besides, in the action shielding mechanism, we pre-execute RL actions with MPC and also provide theoretical proofs demonstrating its effectiveness in enhancing RL's safety and sample efficiency. Moreover, due to the use of safety constraints and action shielding in this study, safety violation behaviors can be significantly reduced during the training process. } The main contributions are summarized as follows:
\begin{itemize}
    \item {A human-aligned safe RL method based on CMDP is proposed for the highway on-ramp merging task, in which the individual's risk preference is incorporated into the safety constraints to accommodate the safety expectations of users. The safety level of the RL policy can be adjusted by determining the cost limits of constraints of CMDP using a fuzzy decision-making method based on the inputs of user preferences and traffic density.}
    \item {An ASM is built to mask out unsafe RL actions. We pre-execute the RL action with MPC and conduct collision checks with surrounding agents to determine whether the action is safe. Theoretical proof has shown the effectiveness of the shielding mechanism in terms of safety and sampling efficiency.} 
    \item {Numerical simulations in different levels of traffic densities have shown that our method outperforms the baselines, which can improve safety without sacrificing traffic efficiency. Due to the use of safety constraints and action shielding, risk behaviors can be significantly reduced during the exploration stage of RL, providing a possible solution for online training in real-world conditions.} 
\end{itemize}

\subsection{Organization of the paper}
The rest of this paper is organized as follows. Section \ref{relatedwork} provides a comprehensive overview of the related works in the field, laying the groundwork for understanding the current research landscape. Section \ref{problem_formulation} describes the problem statements in detail. Section \ref{MPC} delves into the specific implementation process of MPC, offering a step-by-step explanation of the methodology employed. Section \ref{safe RL} introduces the human-aligned RL algorithm and the action shielding mechanism. Section \ref{theoretical analysis} presents a theoretical analysis of the shielding mechanism regarding safety performance and convergence rate. The detailed implementation process, experimental results, and an in-depth discussion of the findings are presented in Section \ref{Implementation Details} and \ref{results}, respectively. Section \ref{conclusion} concludes the paper.

%%% Literature review
\section{Related Works}\label{relatedwork}
\subsection{{MPC-based Approach}}
{With its remarkable capacity for handling multiple constraints, MPC inherently possesses an advantage in addressing complex decision-making and planning problems. In \cite{9913938}, a combined decision-making and trajectory planning framework is proposed for collision avoidance. Employing the Big-M method, it translates decision-making into inequality constraints, which are subsequently integrated into the constraints of MPC. In \cite{wang2024high}, an AV is abstracted as a switched system, where the overtaking task is formulated as a mixed-integer programming (MIP) problem. Similarly, \cite{8619852} represents vehicles on highways as a mixed-logical-dynamical (MLD) system and formulates the coordination problem as a generalized mixed-integer potential game. In \cite{8289418}, a probabilistic and deterministic prediction-based lane change decision-making method is proposed, where the stochastic MPC (SMPC) with safe driving envelope constraints is formulated to generate control input decisions. However, safety constraints in SMPC are expressed as probabilistic constraints, which cannot guarantee complete safety. To address this, \cite{9410387} proposes a framework that combines SMPC and fail-safe trajectory planning to obtain safe and efficient trajectories. In \cite{8814171}, an MPC-based lane merging approach is developed, where a hierarchical control system is designed to determine the optimal cut-in location and the appropriate vehicle to follow. \cite{7990647} presents a scenario MPC method for lane change on highways, accounting for the uncertainty in the traffic environment by a small number of future scenarios. \cite{Dixit2020} use potential field function and reachability sets to identify safe zones, which are provided to a tube-based robust MPC as a reference to generate feasible trajectories for combined lateral and longitudinal motion. Overall, MPC seeks the optimal control that maximizes objectives with safety constraints considering the vehicle dynamics and interactions with other traffic participants. The performance of MPC-based approaches is determined by the precision of the system model and the efficiency of the solution \cite{cao2015cooperative}. However, this does not obscure the superiority of MPC in handling complex constraints. In this paper, MPC is utilized in an ASM to filter out unsafe decision actions and motion planning in the lower-level controller.}
%%%  RL
{
\subsection{RL-based Approach}} {RL has shown great potential in the complex decision-making task of autonomous driving  \cite{Wang2017, Lin2020, Wang2021, chen2023, wang2021deep, chen2023research, wang2021tactical, yang2022generalized}.
For instance, a Deep Q-Network is built to learn the optimal vehicle control \cite{Wang2017}, and a Deep Deterministic Policy Gradient method is employed to determine the optimal acceleration of the merging vehicle \cite{Lin2020}. 
The reward function is a crucial part of RL, which guides the learning process, and it is a non-trivial task to design proper reward functions due to the difficulty in balancing the multiple objectives such as safety, comfort, and efficiency \cite{Aradi2022, gu2022review}, etc. Safety is one of the most important goals in safety-critical domains such as autonomous driving \cite{Udatha2023}, and safety violations are not allowed in real-world implementations \cite{achiam2017, yang2021wcsac, yang2023safety, yang2023reinforcement}. Many studies optimize the weighted sum of safety and other objectives \cite{Wang2017, Lin2020, Wang2021}, which assigns different weights to each objective. \cite{Wang2017} build reward functions to measure safety, smoothness, and efficiency, and \cite{Wang2021} formulates the reward function as the combination of efficiency and safety. However, the optimal policy that maximizes the weighted sum of rewards cannot ensure that the safety reward is maximized, and thus, safety violations still happen \cite{achiam2017}.} \par
{To directly optimize the safety terms, CMDP is employed, which formulates the safety violations in the cost function and constrains the expectation of the cumulative costs under a specific safe threshold \cite{achiam2017, yang2021wcsac, Udatha2023}. The primal-dual approaches are widely used to solve the CMDP, in which Lagrange multipliers for the cost constraints are introduced to form the Lagrangian function, and the dual problem can be defined as a Lagrangian relaxation of the primal problem. The primal-dual method typically involves iteratively solving the dual and the primal problem until both the primal and dual solutions converge. However, the iterative updates between the primal and dual variables can cause oscillations, particularly if the learning rates are not appropriately set. The primal-dual approach also becomes computationally intensive when dealing with CMDPs with large state and action spaces. Besides, in non-stationary environments where the dynamics or reward structures change over time, primal-dual methods might struggle to adapt quickly or fall into local optima. Overall, scalability in high-dimensional systems, convergence issues, sensitivity to hyperparameters, and adaptation to non-stationary environments pose the key challenges for primal-dual methods in solving CMDP for safe RL applications.} 
%%% 
{\subsection{Shielding Mechanism} Though CMDPs provide a formal mathematical framework to explicitly incorporate safety constraints into the decision-making process, CMDP-based safe RL methods are difficult to enforce the hard safety constraints, and the agent may still fall into unsafe regions \cite{pmlr-v202-wang23as}. To address the unsafe action generated by the RL agent, the shielding mechanism is designed to monitor and correct the risk action when it violates safety constraints \cite{carr2023safe}. Motion prediction is usually incorporated into the shielding module, allowing the agent to anticipate future states and avoid actions that may result in unsafe states \cite{Isele2018, Krasowski2020, chen2023}. For instance, \cite{chen2023} proposes an action masking scheme to filter out invalid/unsafe
actions, which exploits the vehicle dynamics to predict the potential collisions over a prediction horizon and correct unsafe actions if necessary. In \cite{Isele2018}, a prediction module is built to mask risky actions and the agent can explore the safe space using common RL methods. \cite{Krasowski2020} designs a safety layer that determines safe actions based on set-based predictions considering all possible occupancies of traffic participants. Accurate and robust motion predictions are essential to shielding in safe RL, but it requires a good understanding of the system dynamics, uncertainty representation, and the modeling of complex interactions between the traffic participants \cite{konighofer2022online}. Moreover, shielding assumes that an environment is fully known in advance, and thus it would be hard to formulate an appropriate shielding in an unknown environment \cite{Pranger2021}.} 

{Different from \cite{Isele2018, Krasowski2020, chen2023}, in the ASM, we pre-execute the RL action with MPC to acquire the predictions of the ego vehicle and then conduct collision checks with surrounding agents based on their motion predictions to determine whether the RL action is safe or not. Due to the use of the ASM, the unsafe actions are replaced with safe ones and then sent to the simulation environment, which can help the RL agent quickly learn to act safely. Moreover, many studies neglect theoretical proof of the effectiveness of the action shielding mechanism in RL, and we provide a theoretical justification of why the action shielding mechanism can enhance the safety performance and learning efficiency of RL. }
%%%% 
{\subsection{Combination of RL and MPC}}
{RL is more suitable for scenarios where the system dynamics are complicated. However, RL is hard to generalize in unseen environments. MPC can handle complex constraints, but its performance is limited by the model's accuracy and computational complexity. Both RL and MPC have advantages and limitations, and a natural choice is to combine the two. For instance,
\cite{morinelly2016dual} formulates the dual MPC for the simple case of a discrete linear system with uncertain dynamics by introducing the effect of prediction information under approximate dynamic programming (ADP) and RL framework.  The RL agent is viewed as the MPC algorithm, which computes the actions adaptively by solving the optimal control problem. \cite{williams2017information} introduces an information-theoretic version of MPC that allows for handling complex cost criteria and general nonlinear dynamics. \cite{zanon2019practical}, \cite{martinsen2020combining} use a parametric MPC to estimate the value function of RL. 
\cite{bellegarda2020online} reduces the sample complexity of RL by using MPC,  and \cite{karnchanachari2020practical} uses RL to approximate the value function given only high-level objectives.
\cite{gros2021reinforcement} allows for the implementation of RL techniques in the context of MPC-based policies. 
In \cite{Lubars2021}, the MPC controller works as a supervisor of RL, evaluates the safety of RL actions, and replaces the RL action if it is unsafe. \cite{albarella2023hybrid} also uses RL to choose high-level tactical behaviors and Nonlinear MPC for low-level control. 
\cite{chen2023} formulates a multi-agent RL problem for on-ramp merging in mixed traffic, in which AVs learn to adapt to human-driven vehicles and cooperatively accomplish merging tasks safely. A priority-based safety
supervisor is designed to reduce the computation burden by selecting the vehicle with higher priority for safety checks. The combination of RL and MPC has shown great potential in complex decision-making and planning tasks of autonomous driving.}\par
{Existing studies often consider safety in reward terms \cite{Lubars2021,albarella2023hybrid,chen2023}, while we use costs to constrain safety violations by formulating the decision-making problem as a CMDP. We also incorporate individual risk preferences into the constraint settings of CMDP to meet the users' safety expectations, which have rarely been considered in current publications. In addition, we use MPC to pre-execute the RL action for multi-step prediction for collision checks in the action shielding mechanism and also provide theoretical proofs to demonstrate its effectiveness. Finally, MPC is employed again for the low-level trajectory planning based on the high-level safe RL action. We note that MPC has been used twice in this study, one in the ASM and the other in the low-level motion planning module. Moreover, due to the use of constraints and shielding, we can adjust the safety level of the final policy and avoid safety violations during exploration, proving a promising solution for online training of RL in real-world environments.}
%%%%%%
{\subsection{Human Risk Perception in Decision-making}}
{Risk perception in driving has gained wide scientific interest over the years, in which the differences among individuals' risk perception have a significant impact on the safety design of intelligent vehicles \cite{fyhri2012personality, rhodes2011age}. Many studies use statistical methods to analyze driving risk, among which the most common method is to assess risks by analyzing driving behaviors \cite{chen2024quantifying}. To understand the differences in risk perception between drivers, driving style has been studied, including unsupervised methods (clustering, principal component analysis \cite{eboli2017investigating}, and fuzzy classification \cite{constantinescu2010driving}) and direct detection methods (sharp acceleration, braking, and turning, etc.). The recent works \cite{chen2024quantifying} propose a universal driving risk model, which provides a new way of modeling the differences in driver’s risk perception at a physical level. 
Risk perception is also the main consideration in the decision-making for driving safety. \cite{9575928} uses the risk of each sample trajectory as the cost item and obtains trajectory with minimal cost to guarantee the safety requirement. \cite{10195149} proposes a model to describe the acceptable risk and shows how an accepted risk contributes to the decision-making of AVs at the maneuver level. \cite{10184920} proposes a risk-aware decision-making framework to handle the epistemic uncertainty arising from training the prediction model on insufficient data. }

{Different from previous works, this study considers human risk perception in the high-level RL decision process and incorporates the individuals' risk preference into the constraint settings of CMDP, which is expected to align the user's preference with the high-level safety decisions. The cost limit value for the CMDP is determined using a fuzzy control method based on the fuzzy inputs of user preferences and traffic density, indicating that the safety level of the decision policy can be adjusted according to human risk preference and traffic density.}
%%%%%%
%%% New section
\section{Problem Statement}\label{problem_formulation}
This section formulates the decision-making process of the on-ramp merging problem as a hierarchical framework that consists of a CMDP for high-level maneuver decisions and MPC for low-level motion control.  
%% overview
\subsection{Overview of Framework}
The study considers the highway on-ramp merging scenario for autonomous driving, as depicted in Fig. \ref{fig:merge scenario}, where the ego vehicle is randomly spawned on the ramp and aims to safely and efficiently merge into the dense traffic. We introduce a hierarchical decision-making framework, as shown in Fig. \ref{srl mpc architecture}, in which RL determines the high-level maneuvers (\textit{e.g.}, acceleration, deceleration, or lane changes), and MPC determines low-level control. We formulate the high-level decision problem as CMDP, which considers different risk levels in the safety constraints and solves them with a discrete SAC algorithm. An ASM is built to filter out the unsafe/invalid RL actions based on collision checks with predicted states generated from the MPC module that performs motion prediction of the raw RL action. The unsafe/invalid actions from the RL agent are replaced with safe ones and the safe actions are finally sent to the low-level model predictive controller to determine the optimal vehicle control.
%% CMDP
\subsection{Constrained Markov Decision Process}
CMDPs are an extension of the standard MDP that incorporates constraints into the decision-making process. The high-level maneuver decision is formulated as a CMDP model in this study, which specifies safety requirements as constraint terms. CMDP is described as a tuple $<S, A, P, r, c,\eta,\gamma>$, where $S$ is the state space,  $A$ is the action space, $P$ is the transition probability $p(s_{t+1} | s_t,a_t)$ from state $s_t$ to the next state $s_{t+1}$ under the action $a_t\in A$, $r:S\times A \rightarrow \mathbb{R}$ is the reward function, $c:S\times A \rightarrow \mathbb{R}$ is the cost function.

%% figure 3
\begin{figure*}[t]
	\centering
	\subfloat[]{\includegraphics[width=0.66\columnwidth]{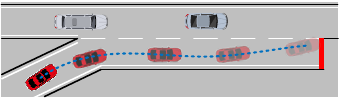}\label{crash with obj}}
	\hfil
	\subfloat[]{\includegraphics[width=0.66\columnwidth]{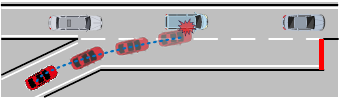}\label{crash with v}}
	\hfil
	\subfloat[]{\includegraphics[width=0.66\columnwidth]{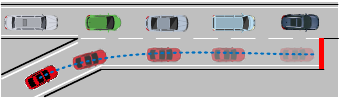}\label{hard to merge}}
     \caption{Cost design based on motion predictions of the ego vehicle and surrounding objects. There are three typical situations, including (a) failing to merge before reaching the end of the road, (b) colliding with other vehicles, and (c) being hard to merge when the target lane is occupied by other vehicles.}
	\label{cost function}
\end{figure*}

%% state
\subsubsection{\textbf{State Space}} let $s^i=(p^i,x^i,y^i,v^i_x,v^i_y)$ be the state of vehicle $i$, where $p^i$ is a binary variable indicating whether the vehicle $i$ is observable, $x^i$ and $y^i$ are longitudinal and lateral distance between the vehicle $i$ and the ego vehicle, $v^i_x$ and $v^i_y$ are relative speed between the vehicle $i$ and the ego vehicle in longitudinal and lateral directions. The complete state of the environment consists of all vehicles that are observed by the ego vehicle, and the state space $S$ is written as,
\begin{equation}
    S = \bigl(s^1,s^2,\cdots,s^n\bigr),
\end{equation}
where $S$ is the state space, $n$ is the maximum number of the observed vehicles on the main lane. 
%% action
\subsubsection{\textbf{Action Space}} the high-level decision module outputs a discrete action, including turning left/right, idling, acceleration/deceleration. The action space $A$ is defined as,
\begin{equation}
\label{action space}
    A = \bigl[A^\triangleleft, A^\triangleright, A^\uparrow, A^\oslash, A^\downarrow\bigr],
\end{equation}
where $A^\triangleleft, A^\triangleright, A^\uparrow, A^\oslash,$ and $A^\downarrow$ are turning left, turning right, acceleration, idling, and deceleration, respectively.
%% reward
\subsubsection{\textbf{Reward}} the reward functions are defined considering safety, success ratio, and traffic efficiency,
\begin{equation}
    r = r_{\text{s}}+r_{\text{g}}+r_{\text{v}},
\end{equation}
where $r$ is the sum of rewards, consisting of the reward of safety $r_{\text{s}}$, the reward of successfully merging and reaching the goal position $r_{\text{g}}$, the reward of traffic efficiency $r_{\text{v}}$. In particular, $r_{\text{s}}$ and $r_{\text{g}}$ are set as constant values, and the reward of traffic efficiency $r_{\text{v}}$ is defined based on ego vehicle speed and traffic speed, which encourages the ego vehicle to drive at the average speed of the traffic flow,
\begin{equation}
\label{velocity reward}
{r_\text{v}} =
\begin{cases}
0.1 & {\text{if}}\ {\left| v_{\text{ego}} - v_{\text{ave}}\right| \leq \kappa v_{\text{ave}}} \\
-0.5 & {\text{else}}
\end{cases},
\end{equation}
where $v_{\text{ego}}$ is the speed of ego vehicle, $v_{\text{ave}}$ is the average speed of traffic, $\kappa$ is the coefficient of the average speed. The agent receives a bonus of 0.1 if the speed difference between the ego vehicle and traffic speed is lower than $\kappa v_{\text{ave}}$; otherwise, it has a penalty of 0.5 for speeding or driving slowly. 

%% cost
\subsubsection{\textbf{Cost}} we use constraints to represent the driving cost caused by collisions, risky behaviors, and unexpected decisions, and the cost functions can be written as,
\begin{equation}
    c=c_{\text{a}}+c_{\text{r}}+c_{\text{u}},
\end{equation}
where $c$ is the sum of costs, including the cost for collision events $c_{\text{a}}$, the cost for risky behaviors $c_{\text{r}}$, and the cost for the unexpected decision $c_{\text{u}}$. We note that the safety cost $c_{\text{r}}$ for the risky behaviors includes three situations, as illustrated in Fig. \ref{cost function}, which are computed based on the predicted states of the ego vehicle and surrounding objects, including a) fail to merge before reaching the end of the road, b) collide with other vehicles, and c) hard to merge when the target lane is occupied. In particular, we use the longitudinal position and relative speed to determine whether the target lane is occupied, 
\begin{equation}
{1^{i}_{oc}} =
\begin{cases}
1 & {\text{if}}\ {\left| v^{i}_{x}\right| \leq 1.5 } \,\,{and}\,\, {\left| x^{i}\right| \leq 5 } \\
0 & {\text{else}}
\end{cases},   
\end{equation}
\begin{equation}
{1_\text{lane\_occupied}} = \mathop{\text{max}}\limits_{i} {\left[{1^{i}_{oc}}\right]},
\end{equation}
where ${1_\text{lane\_occupied}}$ indicates whether the target lane is occupied, $v^i_x$ is the relative speed in the longitudinal direction, $x^i$ is the longitudinal distance between the vehicle $i$ and the ego vehicle. 

%% Problem Formulation
\subsubsection{\textbf{Problem Formulation}}
We formulate the high-level decision problem as a CMDP model, which aims to find an optimal policy that maximizes the cumulative rewards subject to cost constraints. Mathematically, it can be described as,
\begin{align}
\label{cmdp}
\begin{matrix}
\mathop{\text{max}} J_{\pi}^{R} = \mathbb{E}_{\pi}\bigl[{\sum\limits_{t=0}^\infty}\gamma^tr(s_t,a_t)\bigr],\\
\mathop{\text{s.t.}} J_{\pi}^C = \mathbb{E}_{\pi}\bigl[ \sum\limits_{t=0}^{\infty}\gamma^{t}c(s_t,a_t)\bigr] \leq \eta,
\end{matrix}
\end{align}
where $J_{\pi}^{R}$ and $J_{\pi}^{C}$ are the expected sum of rewards and costs following the policy $\pi$, respectively, $\gamma\in[0,1)$ is a discount factor. 
$r(s_t,a_t)$ and $c(s_t,a_t)$ are reward functions and cost functions. $\eta$ is the cost limit that is determined by the risk level and traffic density. Let $\Pi$ be a set of policies, and the action $a_t$ is sampled from a stochastic policy $\pi \in \Pi$, \textit{i.e.},  $a_t \sim \pi(\cdot | s_t)$. We reformulate the constrained optimization problem into a max-min problem via Lagrange multipliers,
\begin{equation}
\label{max-min problem}
\mathop{\text{max}}\limits_{\pi \in \Pi}\ \mathop{\text{min}}\limits_{\lambda\geq0}J_{\pi}^R-\lambda(J_{\pi}^C-\eta),
\end{equation}
where $\lambda$ is the Lagrange multiplier. The Lagrange multiplier increases once the constraints have been violated.

%% safe cost
\subsection{Human-aligned Safety Cost Limits}
{
As discussed previously, the cumulative costs $J_{\pi}^{C}$ are constrained by the cost limit $\eta$ in Equation (\ref{cmdp}), which determines the risk level of the learned policy. Setting an appropriate value for $\eta$ is essential to match the users' preferences for safety.} \par
{In this study, we utilize the fuzzy control method \cite{fuzzycontrol} to determine the value of the cost limit $\eta$ based on the specific risk level and traffic density. Mamdani inference is often used to create a fuzzy logic control system by synthesizing a set of linguistic control rules obtained from experienced human operators \cite{MAMDANI19751}. the Mamdani inference process involves the following steps, \textit{i.e.}, fuzzification, inference, aggregation, and defuzzification.}
{
\begin{itemize}
    \item \textbf{Fuzzification}. Convert the input values to membership values using a set of membership functions. Membership functions represent how each point in the input space is mapped to a membership value (degree of belonging) between 0 and 1. 
    \item \textbf{Rule evaluation}. Apply the fuzzy rules to obtain the fuzzy output. The fuzzy logic rules are typically represented as "IF-THEN" statements, i.e., ``If $x$ is A and $y$ is B, then $z$ is C", where x and y are input variables, A and B are fuzzy sets associated with $x$ and $y$, and C is the fuzzy set for the output variable $z$. This process involves computing the minimum (AND operation) or maximum (OR operation) of the membership values of each rule.
    \item \textbf{Aggregation}. Combine the outputs of all fuzzy rules to form a single fuzzy set. This step involves taking the union of the fuzzy sets resulting from each rule.
    \item \textbf{Defuzzification}. Convert the aggregated fuzzy output back into a crisp value. The most commonly used method for defuzzification is the Centroid Method, which calculates the centroid of the aggregated fuzzy set to obtain a crisp value.
\end{itemize}}\par
{The membership function is a critical step in developing a fuzzy logic system, which can be designed based on expert knowledge or empirical data. Experts can provide insights into how variables should be divided into fuzzy sets and how these sets should be shaped. Statistical analysis of empirical data can also help identify appropriate membership functions. In this study, we define the membership functions based on human knowledge. Risk level and traffic density are selected as the fuzzy inputs, and the cost limit is the fuzzy output.}

\begin{table}[b]
  \centering
  \caption{Fuzzy relations between the cost limit and traffic density, risk level. Cost limit has the linguistic value of small, medium, and large.
    \label{table:fuzzy_control_rule_table}}
    \renewcommand\arraystretch{1.8}
    {
    \begin{tabular}{cc|ccc}
    \toprule[1pt]
          &       & \multicolumn{3}{c}{Traffic Density} \\
          &       & High  & Medium & Low \\
    \midrule[1pt]
    \multirow{3}{*}{Risk Level} & Conservative & Small & Small & Medium \\
          & Neutral & Small & Medium & Large \\
          & Aggressive & Medium & Large & Large \\
    \bottomrule[1pt]
    \end{tabular}%
    }
  \label{tab:addlabel}%
\end{table}%

\begin{figure*}[t]
\centering
\subfloat[]{\includegraphics[width=2.0in,trim=0 0 0 0, clip]{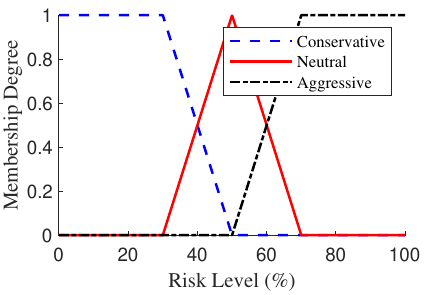}{\label{fig:risk_level_membershipfunc}}}
\hspace{0.1in}
\subfloat[]{\includegraphics[width=2.0in,trim=0 0 0 0, clip]{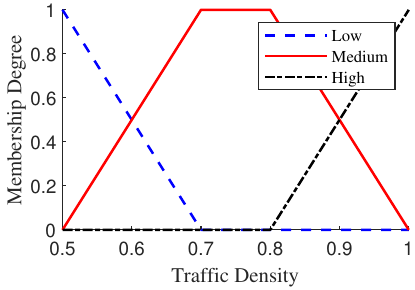}{\label{fig:traffic_density_memberfunc}}}
\hspace{0.1in}
\subfloat[]{\includegraphics[width=2.0in,trim=0 0 0 0, clip]{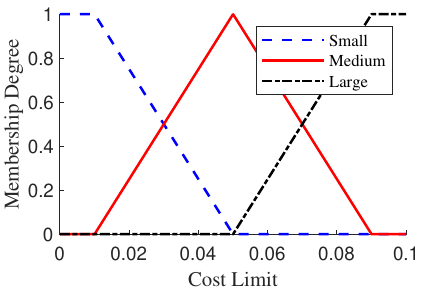}{\label{fig:costlimit_membershipfunc}}}
\caption{{The membership functions of the fuzzy inputs risk level and traffic density, and the fuzzy output cost limit. (a) risk level, varying between 0 and 100\%. (b) traffic density, varying between 0.5 and 1. (c) cost limit, varying between 0 and 0.1.}}
\label{fig:membershipfunc}
\end{figure*}

{The membership functions of risk level, traffic density, and cost limit are shown in Fig. \ref{fig:membershipfunc}. We define three types of risk levels, including \{`conservative', `neutral', `aggressive'\}, as shown in Fig. \ref{fig:risk_level_membershipfunc}. Traffic density includes \{`low', `medium', `high'\}, as shown in Fig. \ref{fig:traffic_density_memberfunc}, and the traffic density varies between 0.5 and 1.0. The membership function of traffic density is built based on the definition of low, medium, and high levels of traffic density in Section \ref{subsubsec:scenario settings}. From Fig. \ref{fig:traffic_density_memberfunc}, the membership value of low traffic density drops to zero when the traffic density is greater than 0.7. As the medium traffic density ranges from 0.5 to 1, the membership function is represented with a trapezoid curve, which has the highest membership value of 1 between a traffic density of 0.7 and 0.8.  
The membership function of high traffic density is increasing when the traffic density is greater than 0.8. Similarly, the membership function is designed in Fig. \ref{fig:costlimit_membershipfunc}, including \{`small', `medium', `large'\}, and the cost limits vary between 0 to 0.1.}

{After fuzzification, we can define fuzzy rules that map the fuzzy inputs (risk level and traffic density) to the fuzzy output (cost limit). Fuzzy rule is a crucial aspect of designing a fuzzy logic system, and the rules are usually derived from expert knowledge or empirical data. Experts in the domain provide the rules based on their understanding and experience, and this approach is often used in fields where human expertise is available. In addition, rules can be derived from empirical data using statistical analysis or machine learning techniques, which is beneficial when expert knowledge is unavailable or an empirical dataset is ready to use. In this study, the fuzzy rules are defined based on human knowledge, and Table \ref{table:fuzzy_control_rule_table} shows the fuzzy relations between the cost limit, and risk level, traffic density. The cost limit has the linguistic values of \{`small', `medium', `large'\}. From Table \ref{table:fuzzy_control_rule_table}, we can infer the fuzzy output of cost limit given the fuzzy inputs of risk level and traffic density. For instance, if traffic density is `low' and the risk level is `aggressive', then the cost limit is `small'. We use a matrix $\tilde{R}$ to represent the fuzzy rules, and the process of rule evaluations can be written as,  }
%% rule evaluation

\begin{equation}
\label{loss_lambda}
    {\tilde{C} = \bigcup\limits_{k=1}^{9}(\tilde{A}_{i} \times \tilde{B}_{j}) \circ \tilde{R}_{k},}
\end{equation}
{where $\tilde{C}$ is the fuzzy output, $\tilde{A}_{i}$ is the traffic density, $\tilde{B}_{j}$ is the risk level, $i,j\in\{1,2,3\}$ are the indices of different fuzzy sets w.r.t. risk level and traffic density, and the operator $\circ$ denotes the composition of fuzzy relations. }

\begin{figure}[t]
    \centering
    \includegraphics[width=3.2in,trim=0 0 0 0, clip]{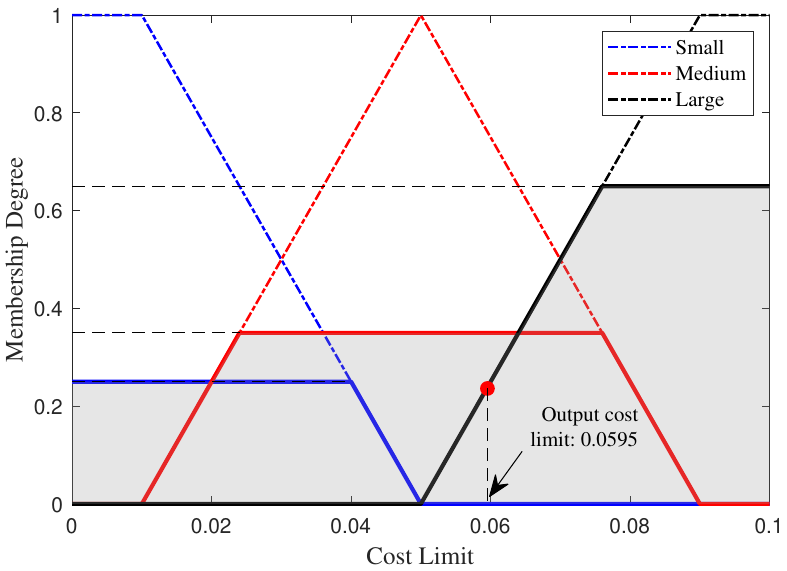}
    \caption{{Illustration of the aggregation and defuzzification of the Mamdani inference process. The fuzzy sets of the cost limit include large, medium, and small, which are represented with black, red, and blue dash lines. Given that the traffic density and risk level are 0.57 and 45$\%$, and the membership value of the input is $\tilde{A} = \{\text{Low}, \text{Medium}\}$ and $\tilde{B} = \{\text{Conservative}, \text{Neutral}\}$, we can compute the fuzzy output for the cost limit of small, medium, and large, which are 0.25, 0.35, and 0.65, respectively. With the fuzzy output, we can compute the grey area for each fuzzy set and aggregate them into a single fuzzy set, which is the union area of three grey areas. Finally, we take the centroid of the grey area to obtain a crisp value for the cost limit, which is 0.0595. That is, with the inputs of traffic density and risk level of 0.57 and 45\%, the cost limit is set as 0.0595.}}
    \label{fig:Mamdani inference}
\end{figure}

{Based on the fuzzy rules, the fuzzy output of each rule can be computed. These fuzzy outputs are aggregated into a single fuzzy set and converted to the crisp output value using the centroid method. Fig. \ref{fig:Mamdani inference} illustrates an example of aggregation and defuzzification of the Mamdani process. Given that the traffic density and risk level are 0.57 and 45$\%$, and the membership value of the input is $\tilde{A} = \{\text{Low}, \text{Medium}\}$ and $\tilde{B} = \{\text{Conservative}, \text{Neutral}\}$, we can compute the fuzzy output for the cost limit of small, medium, and large, as shown in Fig. \ref{fig:Mamdani inference}, which are 0.25, 0.35, and 0.65 respectively. With the fuzzy output, we can compute the grey area for each fuzzy set and aggregate them into a single fuzzy set, which is the union area of three grey areas, as shown in Fig. \ref{fig:Mamdani inference}. Finally, we take the centroid of the unions of grey areas to obtain a crisp value for the cost limit, which is 0.0595. Therefore, the cost limit is 0.0595 when traffic density is 0.57 and risk level is 45\%. }

%% Fig, fuzzy control

%% Fig, Mamdani inference result

%%%%%%%%% MPC
\section{Model Predictive Control}
\label{MPC}
MPC is an optimization-based approach that predicts the future states of a system over a finite time horizon and computes the optimal control input that minimizes the cost function while satisfying a set of constraints. MPC requires a dynamic model that represents how the system evolves in response to control inputs, a cost function that quantifies the system's performance, and constraints that limit the state variables and control inputs. The optimization process is solved at each time step, and the first control action from the optimized sequence is used. In our study, MPC has been used twice; one is in the ASM, and the other is in the low-level controller. As discussed previously, MPC first executes the RL action and predicts the future states of the ego vehicle over a finite time horizon. Based on the predicted states, the ASM can detect unsafe actions and replace them with safe ones. With the safe one provided by the ASM, the MPC controller can finally compute the optimal control to maneuver the vehicle. 
\subsection{Discrete Linear Model}
\begin{figure}[!t]
\centering
\includegraphics[width=2.5in]{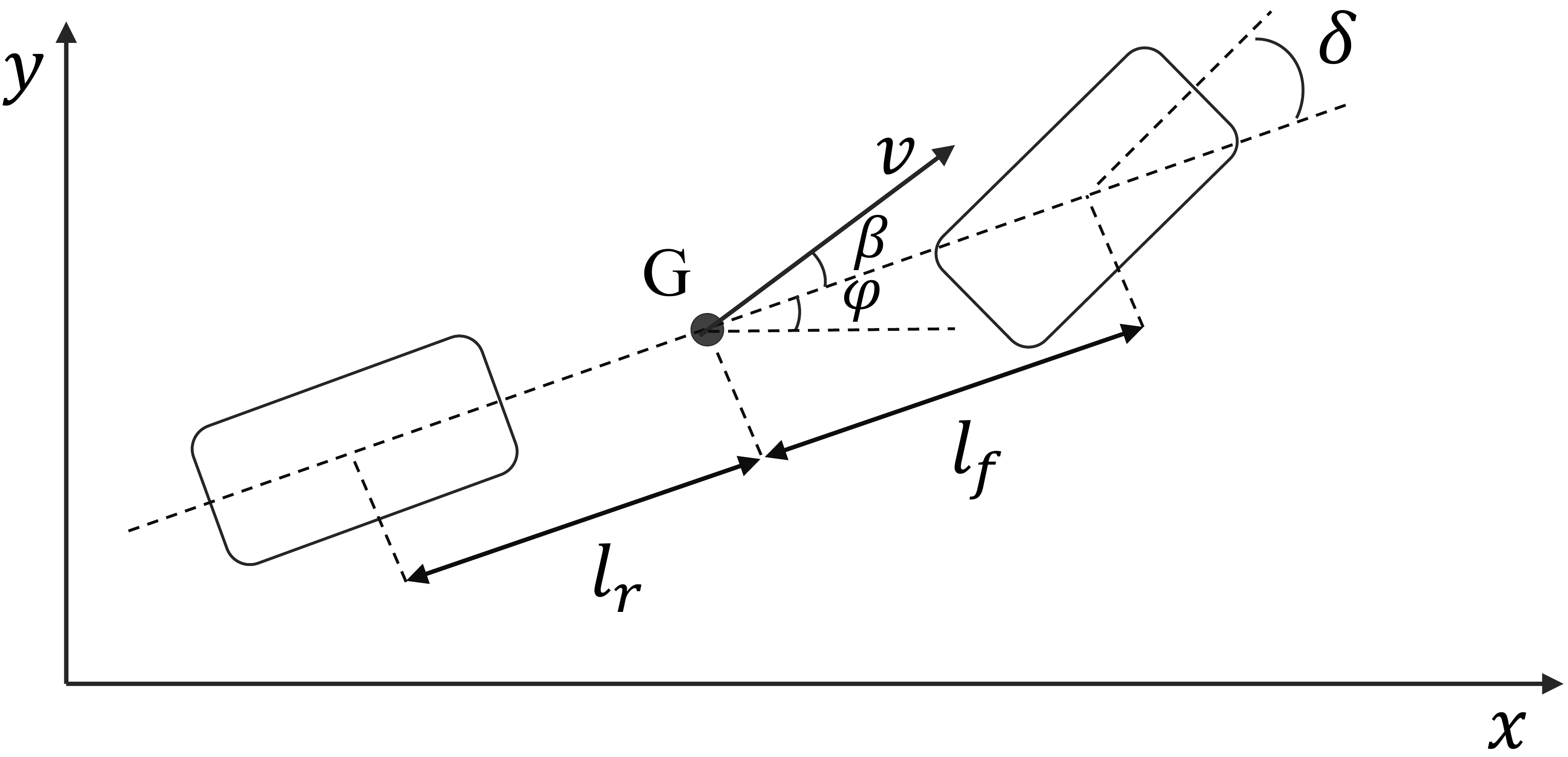}
\caption{Kinematic bicycle model.}
\label{kinematic model}
\end{figure}

In this study, we use the kinematic bicycle model in MPC \cite{9843863}. As illustrated in Fig. \ref{kinematic model}, the vehicle state is written as $\bold{X}=[x, y, v, {\varphi}]^{\top}$, where $x$ and $y$ denote the longitudinal and lateral position, $v$ and $\varphi$ are vehicle speed and the yaw angle. The side-slip angle, wheelbase, and length from the front and rear axles to the center of gravity are denoted as $\beta$, $l$, $l_f$, and $l_r$, respectively. The control variable is denoted as $\bold{U}=[u_1, u_2]^\top$, where $u_1$ and $u_2$ are acceleration $a$ and the steering angle $\delta$, respectively. The referenced position and control variable are defined as $[x_r, y_r, v_r, {\varphi}_r]^{\top}$ and $[v_r, \delta_r]^{\top}$. We derive a linear model,
% linear model
\begin{align}
    % x_dot = f(x,u)
    \dot{\bold{X}} = f(\bold{X},\bold{U}) = \left[ {\begin{array}{*{20}{c}}{v\cos(\varphi  + \beta )}\\{v\sin(\varphi  + \beta ))}\\a\\{\frac{v}{l}\sin \beta }\end{array}} \right],\\
    % beta = arctan(1/2*tan(delta))
    \beta=\arctan(\frac{l_r}{l_f+l_r}tan\delta),\\
    % ex_dot = Aex + Beu
    \bold{\dot{e}}_\bold{X} = A \bold{e}_\bold{X} + B \bold{e}_\bold{U},\\  
    % A = []
    A=\left[ {\begin{array}{*{20}{c}}
    0&0&{\cos ({\varphi _r} + {\beta _r})}&{ - v_r\sin ({\varphi _r} + {\beta _r})}\\
    0&0&{\sin ({\varphi _r} + {\beta _r})}&{v_r\cos ({\varphi _r} + {\beta _r})}\\
    0&0&0&0\\
    0&0&{\frac{{\sin {\beta _r}}}{l}}&0
    \end{array}} \right],\\
    B=\left[ {\begin{array}{*{20}{c}}
    0&{ \frac{- {v_r}\sin ({\varphi _r} + {\beta _r})l_r(l_f+l_r)}{(l_f+l_r)^2\cos^2{\delta_r}+l_r\tan^2{\delta_r}}}\\
    0&{ \frac{- {v_r}\cos ({\varphi _r} + {\beta _r})l_r(l_f+l_r)}{(l_f+l_r)^2\cos^2{\delta_r}+l_r\tan^2{\delta_r}}}\\
    1&0\\
    0&{ \frac{\frac{{{v_r}}}{l}\cos \beta_rl_r(l_f+l_r)}{(l_f+l_r)^2\cos^2{\delta_r}+l_r\tan^2{\delta_r}}}
    \end{array}} \right],
\end{align}
where $A \in \mathbb{R}^{4 \times 4}$ and $B \in \mathbb{R}^{4 \times 2}$ are the Jacobi matrix of the function $f$ with respect to state $\bold{X}$ and control $\bold{U}$, respectively. The state error between the predicted state and the reference state is denoted as $\bold{e}_\bold{X} \in \mathbb{R}^{4}$. The control error between the predicted control and the reference control is denoted as $\bold{e}_\bold{U} \in \mathbb{R}^{2}$. We discrete the linear model,
% discrete model
\begin{align}
\label{eq: discrete dynamics}
    \bold{e}_\bold{X}(k) = A_{k}\bold{e}_\bold{X}(k-1) + B_{k}\bold{e}_\bold{U}(k-1),
\end{align}
where $\bold{e}_\bold{X}(k)$ is the state error at time $k$, $\bold{e}_\bold{X}(k-1)$ is the state error at time $k-1$, and $\bold{e}_\bold{U}(k-1)$ is the control error at time $k-1$. The system matrix $A_{k}$ and input matrix $B_{k}$ are defined as
\begin{align}
    A_{k} = A \cdot \Delta t + \text{I}, \\
    B_{k} =  B \cdot \Delta t,
\end{align}
where $\Delta t $ is the time interval. We define the predicted state error vector $E_\bold{X}$ and the control error vector $E_\bold{U}$ as,
\begin{align}
    E_\bold{X} & = [\bold{e}_{\bold{X}}^{\top}(k+1),...,\bold{e}_{\bold{X}}^{\top}(k+N)]^{\top}, \\
    E_\bold{U} & = [\bold{e}_{\bold{U}}^{\top}(k),...,\bold{e}_{\bold{U}}^{\top}(k+N-1)]^{\top},
\end{align}
where $N$ is the predicted time horizon. Then, the optimization problem can be formulated as, 
\begin{align}
\label{optimization problem}
\begin{matrix}
    \text{min} &  E_\bold{X}^{\top}QE_\bold{X} + E_\bold{U}^{\top}PE_\bold{U}, \vspace{1.25ex}
\end{matrix}
\end{align}
where $Q$ and $P$ are the diagonal matrices representing the weight of the state and control errors respectively. The constraints on the control variables are described as,
\begin{align}
\label{constraints}
\begin{matrix}
    & -\tilde{a}_{\text{max}}{\le} u_{1}(k+i) {\le}\tilde{a}_{\text{max}}, i=0,1,...,N-1 \vspace{1.25ex}, \\ 
    & -\delta_{\text{max}}{\le} u_{2}(k+i) {\le}\delta_{\text{max}},i=0,1,...,N-1,
\end{matrix}
\end{align}
where the max acceleration $\tilde{a}_{\text{max}}$ and the max steering angle $\delta_{\text{max}}$ are the upper bound of the control variables, which are usually designed to ensure the driving comfort and tracking performance. This optimization problem is a quadratic programming (QP) problem that can be quickly solved by a nonlinear programming solver, \textit{e.g.}, CVXOPT. The quadratic program solver is denoted as `QP.solver' at line 8 (see Algorithm \ref{alg:mpc}) to solve the optimization problem defined in Equation \ref{optimization problem}. The solution of this optimization problem is $[\bold{e}_{\bold{U}}^{\top}(k),...,\bold{e}_{\bold{U}}^{\top}(k+N-1)]^{\top}$, and we take the first one  $\bold{e}_\bold{U}^{\top}(k)$ as the optimal control.
%% states computation
\subsection{States Computation}
In this study, we compute the state error $\bold{e}_\bold{X}(k)$ at each time $k$ based on the predicted positions for the entire time horizon, as shown in Algorithm \ref{alg:mpc}. We first need to compute reference positions in the Frenet frame to acquire the predicted positions and then convert them back to the Cartesian coordinates. The reference positions are the waypoints along the reference line in the Frenet frame \cite{werling2010optimal}. In the first step, the reference position $s_\text{r}$ in the Frenet frame is derived by $x_\text{r}$, which is the orthogonal projection point of $x$ relative to the reference path. In the next step, the relationship between $x_r$ and $s_r$ is prior-known. For example, if the reference path is a long straight road, then the relationship can be expressed as $x_r = s_r$. It is assumed that the reference line is the center line of the target lane. The target lane is located on the left lane if the decision is to turn left, and vice versa. Therefore, the reference positions in the Frenet frame can be computed as,
\begin{equation}
\label{compute sr}
s_r(k) = s_r(k-1) + v_r\Delta t,
\end{equation}
where $s_\text{r}(k)$ is the reference position at time $k$, $v_r$ is the reference speed that is determined by the decision $a_t$, which is updated as,
\begin{equation}
\label{eq:compute vr}
    v_r=
    \begin{cases}
    v_t + \Delta v & \text{if} \ a_t = A^\uparrow \\
    v_t - \Delta v & \text{if} \ a_t = A^\downarrow \\
    v_t & \text{else}
    \end{cases},
\end{equation}
where $v_{t}$ is the current speed of ego vehicle, $a_t$ is the high-level decision determined by the RL agent and the action space $A$ is defined as, \textit{i.e.}, $A=[A^\triangleleft, A^\triangleright, A^\uparrow, A^\oslash, A^\downarrow]$, which are turning left, turning right, acceleration, idling and deceleration, respectively. In particular, we execute the RL decision in MPC for motion prediction of ego vehicle, and the predicted position of ego vehicle $x^{\sigma}_e$ is written as,
\begin{equation}
\label{compute xp}
x^{\sigma}_e = x_r(\sigma) + e^{x}_{X}(\sigma),
\end{equation}
where $\sigma \in [1,N]$ is the prediction coefficient, $x_r(\sigma)$ is the reference longitudinal position in the Cartesian coordinates and $e^{x}_{X}(\sigma)$ is the longitudinal position error computed by Eq. (\ref{eq: discrete dynamics}).

%% algorithm
\begin{algorithm}
\caption{MPC and States Prediction}\label{alg:mpc}
\renewcommand{\algorithmicrequire}{\textbf{Input:}}
\renewcommand{\algorithmicensure}{\textbf{Output:}}
\begin{algorithmic}[1]
\REQUIRE $s_t$, $a_{t}$, $a_{t}^{\odot}, $ QP.solver.  MPC is used for state predictions (MPC.predict) and executing control (MPC.execute).
\IF{MPC.predict}    
    \STATE $v_r = g(a_t)$ by (\ref{eq:compute vr})      \COMMENT{execute the raw action} 
\ELSIF{MPC.execute}    
    \STATE $v_{r} = g(a_{t}^{\odot})$ by (\ref{eq:compute vr})   \COMMENT{execute the replaced action}  
\ENDIF
    \STATE Get $s_r$ by (\ref{compute sr})    \COMMENT{reference positions}
    \STATE $(x_r,y_r) \leftarrow (s_r, l_r)$ by coordinate transform
    \STATE $\bold{U}^{*} \leftarrow $ QP.solver$(x_r, y_r, v_r)$   \COMMENT{solve the optimal control}
    \STATE Get $\bold{e}_{\bold{X}}$ by (\ref{eq: discrete dynamics})  \COMMENT{state errors}
    \STATE Get $x_{e}^{\sigma}$ by (\ref{compute xp})   \COMMENT{predicted states} 
\IF{MPC.predict}
    \STATE Return $x_{e}^{\sigma}$
\ELSIF{MPC.execute}
    \STATE Return $\bold{U}^{*}$
\ENDIF
\ENSURE $x_{e}^{\sigma}$ or $\bold{U}^{*}$ \COMMENT{predicted states or optimal control}
\end{algorithmic}
\end{algorithm}

%% table: glossary of methodology
\begin{table}[t]
\caption{{The glossary of variables in RL algorithm.} 
\label{table:RL_variables_table}}
\renewcommand\arraystretch{1.25}
\centering
\setlength{\tabcolsep}{2.5mm}
{
    \begin{tabular}{ c c }
    \bottomrule[1pt]
    \makebox[0.23\textwidth][c]{Variable} & \makebox[0.1\textwidth][c]{Meaning} \\
    \bottomrule[1pt]
    $q_{\omega}$ & Soft Q-function\\
    ${q}_{\omega_{c}}$ & Cost value function \\
    $q_{\bar{\omega}_{c}}$ & Target cost function \\
    $Q_{\omega}$ & State-action value \\
    $Q_{\omega_{c}}$ & Soft cost value \\
    $\bar{Q}$ & Target Q function \\
    $\bar{Q}_{c}$ & Target cost value \\
    $\omega$ & Parameter of soft Q function \\
    $\bar{\omega}$ & Parameter of target Q network \\
    $\omega_{c}$ & Parameter of cost network \\
    $\bar{\omega}_{c}$ & Parameter of target cost network \\
    $\pi$ & Policy \\
    $\theta$ & Parameter of policy network \\
    $V$ & State-value function \\
    $\xi$ & Temperature parameter \\
    $J_{\pi}$ & Loss function of policy network \\
    $J_{q}$ & Loss function of soft Q function \\
    $J_{q_{c}}$ & Loss function of cost network \\
    $\lambda$ & Lagrangian multiplier \\
    $\mathcal{D}$ & Replay buffer \\
    \bottomrule[1pt]
    \end{tabular}
}
\end{table}

%%%%%%%%%% safe RL
\section{Safe Reinforcement Learning}
\label{safe RL}
This section introduces the safe RL algorithm, \textit{i.e.}, in which we use a discrete version of the SAC algorithm to solve the CMDP problem. Also, the ASM is built to check the safety of the RL action and replace it when necessary based on the predicted states provided by the MPC module. 
%% D-SAC
\subsection{Lagrangian-based Discrete SAC}
% Figure
\begin{figure*}[t]
    \centering
    \includegraphics[width=6.5in, trim=0 0 0 0,clip]{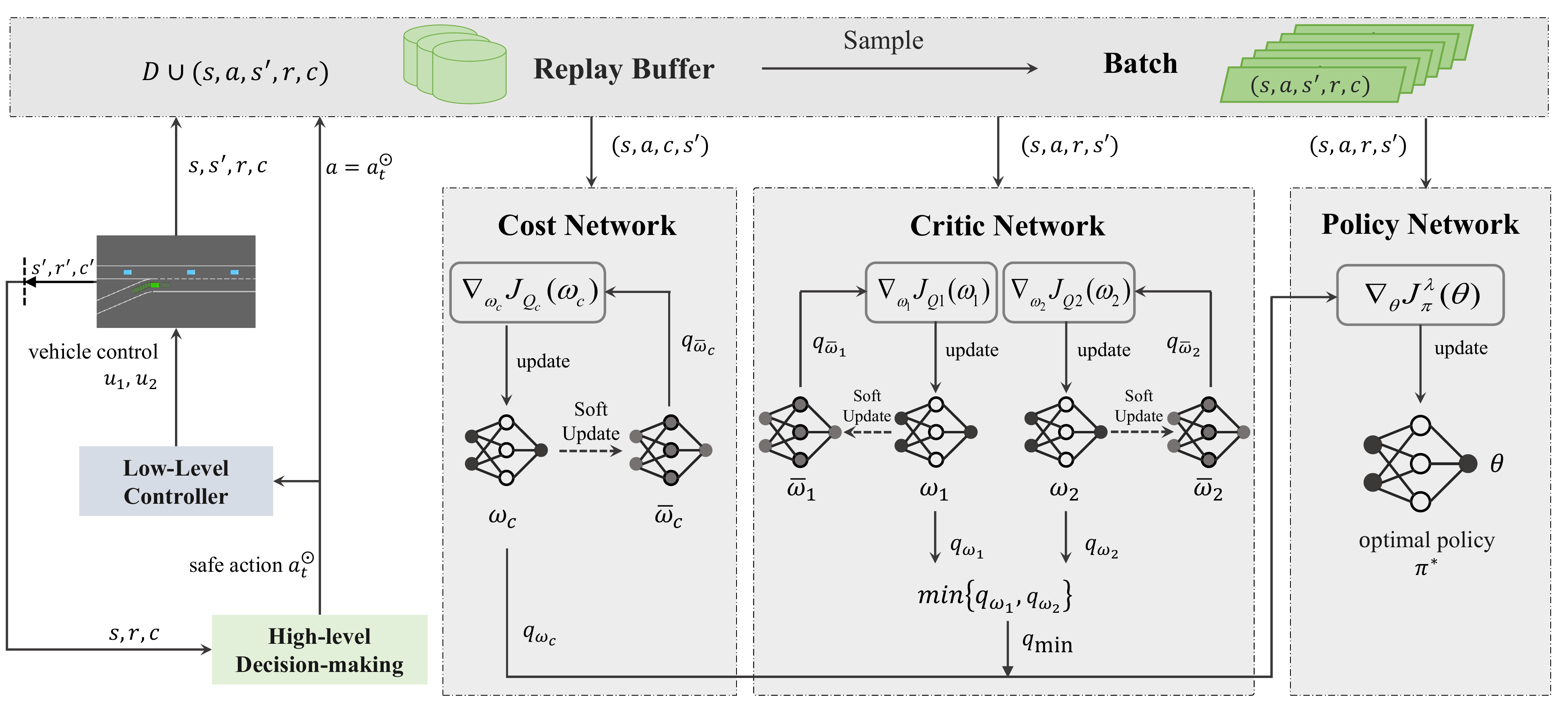}
    \caption{The structure of the neural networks}
    \label{fig:nerual_network}
\end{figure*}
Safe RL is employed for the high-level behavioral decision in this study, \textit{i.e.}, to determine the optimal driving decision from a discrete action space $A$. A discrete action version of the SAC algorithm, \textit{i.e.}, SAC-Discrete (SACD) \cite{christodoulou2019soft}, is used to solve the CMDP problem. \par
\subsubsection{\textbf{Critic Network and Policy Network}}
Let $\mathring{a}^{1},\mathring{a}^{2},..., \mathring{a}^{|A|}$ be the discrete actions in the action space, \textit{i.e.}, $A=\left\{\mathring{a}^{h} \right\}^{|A|}_{h=1}$, $|A|$ is the size of the action space. A critic network and policy network are used in this study, as shown in Fig. \ref{fig:nerual_network}. The soft Q-function of the discrete SAC outputs a vector of size $|A|$ that consists of the Q-value of each action, \textit{i.e.}, $q_{\omega}: S \rightarrow \mathbb{R}^{|A|}$, 
\begin{equation}
   q_{\omega}(s_t)=[Q_{\omega}(s_t,\mathring{a}^{1}),Q_{\omega}(s_t,\mathring{a}^{2}),...,Q_{\omega}(s_t,\mathring{a}^{|A|})]^{T},
\end{equation}
where $q_{\omega}(s_t)$ is the soft Q-function parameterized by
$\omega$, and $Q_{\omega}(s_t,\mathring{a}^{h})$ is the state-action value at state $s_t$ with the discrete action $\mathring{a}^{h}(h=1,2,...,|A|)$.
Likewise, the policy can directly output the action distribution, which outputs a vector that consists of the probability of each action at state $s_t$, \textit{i.e.}, $\pi_{\theta}: S \rightarrow [0,1]^{|A|}$,
\begin{equation}
   \pi_{\theta}(s_t)=[\pi_{\theta}(\mathring{a}^{1}|s_t),\pi_{\theta}(\mathring{a}^{2}|s_t),...,\pi_{\theta}(\mathring{a}^{|A|}|s_t)]^{T},
\end{equation}
where $\theta$ denote the policy network parameters (see Fig. \ref{fig:nerual_network}), $\pi_{\theta}(\mathring{a}^{h}|s_t)$ is the probability of the action $\mathring{a}^{h}(h=1,2,...,|A|)$ conditioned on the state $s_t$. In the discrete action settings, the soft state-value function is computed as,
\begin{equation}
\label{soft state value function}
    V(s_t) = \pi_{\theta}(s_t)^{T}\bigl[{q}_\omega(s_t)-\xi\text{log}\pi_{\theta}(s_t)\bigr].
\end{equation}
where $\xi$ is the temperature parameter that determines the relative importance of the entropy term versus the reward term. According to \cite{christodoulou2019soft}, the policy network parameters $\theta$ can be learned by minimizing the following loss function,
\begin{equation}
    J_{\pi}(\theta) = \mathop{\mathbb{E}}\limits_{s_t\sim\mathcal{D}}\bigl[\pi_{\theta}(s_t)^{\top}\bigl(\xi \text{log}\pi_{\theta}(s_t) - {q}_{\omega}(s_t)\bigr)\bigr].
\end{equation}
The loss function of the temperature parameter is defined as,
\begin{equation}
\label{sacd alpha objective function}
    G(\xi) = \pi_{\theta}(s_t)^{\top}\bigl[-\xi \bigl(\text{log}\pi_{\theta}(s_t) + \bar{\mathcal{H}} \bigr)\bigr],
\end{equation}
where $\bar{\mathcal{H}}$ is a hyperparameter that represents the target entropy. As shown in Fig. \ref{fig:nerual_network}, the critic network parameter $\omega$ of the soft Q-function can be learned by minimizing the soft Bellman residual,
\begin{align}
\label{q loss function}
    J_{q}(\omega) = \mathop{\mathbb{E}}\limits_{s_t,a_t\sim \mathcal{D}}\Bigl[\frac{1}{2}\bigl(Q_{\omega}(s_t,a_t) - \bar{Q}(s_t,a_t)\bigr)^2\Bigr],
\end{align}
\begin{align}
\label{target q}
    \bar{Q}(s_t,a_t) = r(s_t, a_t) + \gamma {\mathbb{E}}_{s_{t+1}}\bigl[V_{\bar{\omega}}(s_{t+1})\bigr],
\end{align}
where $Q_{\omega}(s_t,a_t)$ is the estimated Q function, and $\bar{Q}(s_t,a_t)$ is the target soft Q function. In particular, $V_{\bar{\omega}}(s_{t+1})$ is estimated using a target Q network,  
\begin{equation}
\label{soft state value function}
    V_{\bar{\omega}}(s_{t+1}) = \pi_{\theta}(s_{t+1})^{\top}\bigl[{q}_{\bar{\omega}}(s_{t+1})-\xi\text{log}\pi_{\theta}(s_{t+1})\bigr],
\end{equation}
where $\bar{\omega}$ is the parameter of the target Q network, which is updated in Line 18 in Algorithm \ref{alg:srl with sacd}.

%% Algorithm2
\begin{algorithm}[t]
\caption{Human-aligned safe RL}\label{alg:srl with sacd}
\renewcommand{\algorithmicrequire}{\textbf{Input:}}
\renewcommand{\algorithmicensure}{\textbf{Output:}}
\begin{algorithmic}[1]
\REQUIRE MPC, ASM, $\sigma$, $N$
\REQUIRE initialize $\theta, \omega_{1,2}, \omega_{c}, \bar{\omega}, \bar{\omega}_c, \lambda, \xi, \Gamma$ and $\mathcal{D} \leftarrow \emptyset$
\FOR{each iteration}
    \FOR{each environment step}
    \STATE $ a_t\sim\pi_{\theta}(\cdot |s_t)$
    \STATE $x_{e}^{\sigma}\leftarrow \text{MPC.predict}(s_t,a_t)$ by Algorithm \ref{alg:mpc}
    \STATE $a_t^{\odot}\leftarrow \text{ASM}(a_t,x_e^{\sigma},\sigma,N)$ by Algorithm \ref{alg:spm}
    \STATE $s_{t+1}\leftarrow \text{MPC.execute}(s_t,a_t^{\odot})$ by Algorithm \ref{alg:mpc}
    \STATE $r_t\leftarrow r(s_t,a_t^{\odot})$
    \STATE $c_t\leftarrow c(s_t,a_t^{\odot})$
    \STATE $\mathcal{D} \leftarrow  \mathcal{D} \cup \{(s_t,a_t^{\odot},r_t,c_t,s_{t+1}) \}$
    \ENDFOR
    \FOR{each gradient step}
    
    \STATE $\lambda \leftarrow \lambda - \alpha_{\lambda} \nabla_{\lambda}J(\lambda)$

    \STATE $\omega_{i}\leftarrow \omega_{i}-\alpha_{\omega}\nabla_{\omega_{i}}{J}_q(\omega_{i})$ \text{for} $i={1,2}$
    
    \STATE $\omega_c\leftarrow \omega_c-\alpha_{\omega_c}\nabla_{\omega_c}{J}_{q_{c}}(\omega_c)$

    \STATE $\theta\leftarrow \theta - \alpha_{\theta}\nabla_{\theta}{J}_{\pi}^{\lambda}(\theta)$ 
    
    \STATE $\xi\leftarrow \xi -\alpha_{\xi}\nabla_{\xi} {G}(\xi)$
    
    \IF{soft update}
    \STATE $\bar{\omega} \leftarrow \Gamma\omega+(1-\Gamma)\bar{\omega}$
    \STATE $\bar{\omega_c} \leftarrow \Gamma\omega_c+(1-\Gamma)\bar{\omega_c}$
    \ENDIF
    \ENDFOR
\ENDFOR
\ENSURE Optimal policy $\pi^{*}$
\end{algorithmic}
\end{algorithm}

\subsubsection{\textbf{Cost Network}}
a cost network parameterized by $\omega_{c}$ is built to approximate the cost value function ${q}_{\omega_{c}}(s_t)$, as shown in Fig. \ref{fig:nerual_network}. The cost value function ${q}_{\omega_{c}}(s_t)$ is defined as, 
\begin{equation}
   q_{\omega_{c}}(s_t)=[Q_{\omega_{c}}(s_t,\mathring{a}^{1}),Q_{\omega_{c}}(s_t,\mathring{a}^{2}),...,Q_{\omega_{c}}(s_t,\mathring{a}^{|A|})]^{T},
\end{equation}
where $Q_{\omega_{c}}(s_t,\mathring{a}^{h})$ is the soft cost value at state $s_t$ with the discrete action $\mathring{a}^{h}(h=1,2,...,|A|)$. Similarly, the loss function of the cost network is defined as, 
\begin{align}
\label{cost loss function}
    J_{q_{c}}(\omega_{c}) = \mathop{\mathbb{E}}\limits_{s_t,a_t\sim \mathcal{D}}\bigl[\frac{1}{2}\bigl(Q_{\omega_{c}}(s_t,a_t) - \bar{Q}_{c}(s_t,a_t)\bigr)^2\bigr],
\end{align}
where $Q_{\omega_{c}}(s_t,a_t)$ is the cost value at state $s_t$ and action $a_t$ and $\bar{Q}_{c}(s_t,a_t)$ is the target cost value,
\begin{align}
\label{target cost}
    \bar{Q}_{c}(s_t,a_t) & = c(s_t, a_t) + \gamma {\mathbb{E}}_{s_{t+1}}[V_{\bar{\omega}_{c}}(s_{t+1})],
\end{align}
\begin{align}
    V_{\bar{\omega}_{c}}(s_{t+1}) & = \pi_{\theta}(s_{t+1})^{\top}{q}_{\bar{\omega}_{c}}(s_{t+1}),
\end{align}
where the target cost function $q_{\bar{\omega}_{c}}$ is approximated by the target cost network with parameter ${\bar{\omega}_{c}}$, which is updated in Line 19 in Algorithm \ref{alg:srl with sacd}

\subsubsection{\textbf{$n$-step TD learning}} In traditional temporal difference (TD) learning, the agent updates the value estimates based on the current estimate and a target value. n-step TD learning generalizes this idea by considering not just the immediate next state but by looking ahead n steps into the future. This means that the update is based on the sum of rewards over the next n steps. In this study, we use $n$-step TD learning, which is expected to strike a balance between the short-term and the longer-term perspective of Monte Carlo methods. By adjusting the parameter $n$, the trade-off between bias and variance can be adjusted in the learning process. We also use $\text{n}$-step transitions to approximate the target Q value and cost value functions. According to \cite{tesauro1995temporal}, the target Q-value in Equation \ref{target q} and target cost value in Equation \ref{target cost} are rewritten as,
\begin{align}
\label{multistep baseline Q value}
    \bar{Q}(s_t,a_t) = \sum_{j=0}^{\text{n-1}}{\gamma}^{j}r(s_{t+j}, a_{t+j}) + \gamma^{n} {\mathbb{E}}_{s_{t+1}}[V_{\bar{\omega}}(s_{t+n})], \\
    \bar{Q}_{c}(s_t,a_t) = \sum_{j=0}^{\text{n-1}}{\gamma}^{j}c(s_{t+j}, a_{t+j}) + \gamma^{n} {\mathbb{E}}_{s_{t+1}}[V_{\bar{\omega_{c}}}(s_{t+n})],
\end{align}
where $n$ is the parameter that determines the number of steps that we want to look ahead before updating the Q-function. 
\subsubsection{\textbf{Lagrange multiplier}} we have a cost network, the loss function of the policy network is modified by adding the cost item via the Lagrange multiplier, 

%% J_{\pi}^{\lambda}
\begin{equation}
\label{sacd lagrangian objective function}
J_{\pi}^{\lambda}(\theta)= J_{\pi}(\theta) + \mathop{\mathbb{E}}\limits_{s_t\sim\mathcal{D}}\bigl[\lambda\pi_{\theta}(s_t)^{T} {q}_{\omega_{c}}(s_t)\bigr], 
\end{equation}
where $\lambda$ is the Lagrange multiplier. 
\begin{equation}
\label{gradient-sacd}
\nabla_{\theta} J_{\pi}^{\lambda}(\theta) = \frac{\partial J_{\pi}(\theta)}{\partial \theta} + \lambda \frac{\partial \pi_{\theta}(s_t)}{\partial \theta}^{T} {q}_{\omega_{c}}(s_t).
\end{equation}
%% J(\lambda)
We use the state-action value to approximate the $J_{\pi}^{C}$,
\begin{equation}
\label{loss_lambda}
    J_{\pi}^{C} = Q_{\omega_{c}}(s_t,a_t)
\end{equation}
The loss function of the Lagrange multiplier is denoted as,
\begin{equation}
\label{loss_lambda}
    J(\lambda) = \mathop{\mathbb{E}}\limits_{s_t\sim\mathcal{D}}\bigl[-\lambda (Q_{\omega_{c}}(s_t,a_t)-\eta) \bigr],
\end{equation}
where $J(\lambda)$ is the loss for Lagrange multiplier $\lambda$. \par
Overall, the proposed human-aligned safe RL is described in Algorithm \ref{alg:srl with sacd}. The raw action $a_t$ sampled from the policy $\pi_{\theta}(\cdot |s_t)$ (Line 3) would first be sent to MPC for motion prediction (MPC.predict) (Line 4) and the ASM for safety check (Line 5). The replaced action $a_t^{\odot}$ provided by the ASM is then executed in the low-level MPC controller (MPC.execute, Line 6) and then gets the corresponding cost and reward training (Line 7-8). We use two critic networks parameterized by $\omega_{i}(i=1,2)$ to avoid the overestimation of the soft Q-value \cite{9718218}. In addition, the target soft Q value and cost value are updated based on the parameters $\bar{\omega}$ and $\bar{\omega_c}$.

%% Action Shielding Module
\subsection{Action Shielding Module}
The frequent constraint violation behavior usually makes the RL agent hard to learn. In this study, to enhance the sample efficiency for safe RL, we design a rule-based ASM to replace unexpected or unsafe RL actions with safe ones (see Algorithm \ref{alg:spm}), which can help guide the RL agent to take the right action during the exploration process. In the ASM, the unsafe RL action is detected by checking whether there exist collisions between the ego vehicle and surrounding vehicles based on the predicted states. 

% Algorithm if ASM
\begin{algorithm}
\caption{Action Shielding Module}\label{alg:spm}
\renewcommand{\algorithmicrequire}{\textbf{Input:}}
\renewcommand{\algorithmicensure}{\textbf{Output:}}
\begin{algorithmic}[1]
\REQUIRE $a_t$, $x^{\sigma}_e$, $x_{\text{obj}}$, $\sigma$, $N$
\IF{$a_t$ is $A^{\triangleleft}$}
    \FOR{each vehicle $i$}
    \STATE {$x_{i}(N) = x_{i}(0) + N * v_{i}\Delta t$}
    \IF{$ x_{i}(0) \leq x_{e}^{\sigma} \leq  x_{i}(N)$ and $|y_{e}^{\sigma}(A^{\triangleleft}) - y_{i}| \leq d_y$}
        \STATE $a_{t}^{\odot} = A^{\downarrow}$ \COMMENT{Situation 1}
    \ENDIF
    \ENDFOR
\ENDIF
\IF{$a_t$ is $A^{\triangleright}$}
    \IF{ego vehicle has merged}
        \STATE $a_{t}^{\odot} = A^{\oslash}$  \COMMENT{Situation 2}
    \ENDIF
\ENDIF
\IF{$a_t$ is $A^{\uparrow}$ or $A^{\oslash}$} 
    \IF{$|x_{e}^{\sigma}(a)- x_{\text{obj}}| \leq d_x$}
        \STATE $a_{t}^{\odot} = A^{\downarrow}$ \COMMENT{Situation 3}
    \ENDIF
\ENDIF
% \STATE {Get $a_t^{\odot}$ by Figure \ref{asm_module}}
\ENSURE Safe decision $a_{t}^{\odot}$
\end{algorithmic}
\end{algorithm}

% Figure
\begin{figure}[b]
\centering
\includegraphics[width=3.5in, trim=0 0 0 0, clip]{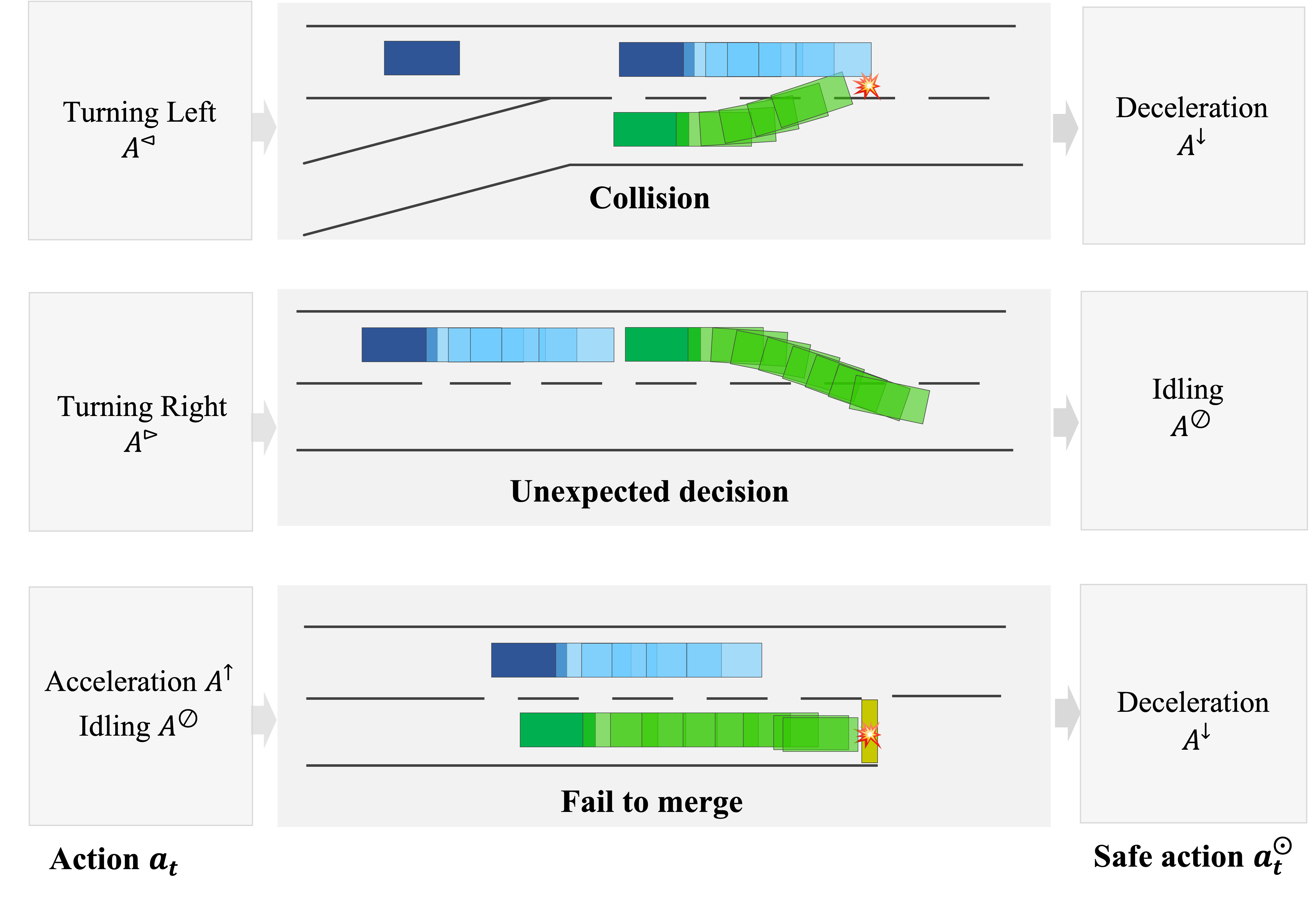}
\caption{Action shielding module. Three typical situations are defined: \textit{i.e.}, collision, unexpected decision, and failing to merge. The unsafe/unexpected action $a_{t}$ (left) is substituted with a safe one $a_{t}^{\odot}$ (right). }
\label{asm_module}
\end{figure}

We define the unsafe decision set as $\Omega_{\text{unsafe}} = \Omega_{1} \cup \Omega_{2} \cup \Omega_{3}$, including three typical situations, as shown in Fig. \ref{asm_module}. If the original RL decisions $a_t \in \Omega_{\text{unsafe}}$, the decisions $a_{t}$ will be substituted with a safe one $a_{t}^{\odot}$.
%% Situation 1
\subsubsection{Situation 1} A collision is likely to occur when the action turning left $A^{\triangleleft}$ is executed (see top Fig. \ref{asm_module}). The unsafe set $\Omega_{1}$ of this situation is defined as,
\begin{equation}
    \Omega_{1} = \{a \in A \vert, x_{i}(0) \leq x_{e}^{\sigma}(A^{\triangleleft}) \leq  x_{i}(N),
    |y_{e}^{\sigma}(A^{\triangleleft}) - y_{i}| \leq d_y\},
\end{equation}
where $x_{e}^{\sigma}$ and $y_{e}^{\sigma}$ are the predicted position of ego vehicle based on the action $A^{\triangleleft}$, $\sigma$ is the prediction coefficient, $x_i$ and $y_i$ are predicted positions of $i^{\text{th}}$ vehicle, and $d_y$ is the safe distance between the ego vehicle and the $i^{\text{th}}$ vehicle in the lateral direction. To ensure safety, the action $A^{\triangleleft}$ is replaced with the deceleration $A^{\downarrow}$, which helps slow down the ego vehicle. The replaced action $a_{t}^{\odot}$ is written as,
\begin{equation}
    a_{t}^{\odot} = A^{\downarrow}.
\end{equation}
%% Situation 2
\subsubsection{Situation 2} The action is labeled as an unexpected decision when the ego vehicle has already merged into the target lane, but the RL agent outputs an action turning right $A^{\triangleright}$, which is considered to be unnecessary (middle in Fig. \ref{asm_module}). The unsafe set $\Omega_{2}$ of this situation is defined as,
\begin{equation}
    \Omega_2 = \{a \in A \vert a = A^{\triangleright}, I_{\text{merge}} = 1\},
\end{equation}
where $I_{\text{merge}}=1$ indicates that the ego vehicle has successfully merged. Therefore, an action idling $A^{\oslash}$ is used to replace the unexpected action $A^{\triangleright}$. The replaced action $a_{t}^{\odot}$ is written as,
\begin{equation}
    a_{t}^{\odot} = A^{\oslash}.
\end{equation}

%% Situation 3
\subsubsection{Situation 3} The ego vehicle's and its neighbors' motion have few differences in the longitudinal direction. The ego vehicle would fail to merge before reaching the end of the merge zone if acceleration $A^{\uparrow}$ or idling $A^{\oslash}$ is executed (see bottom of Fig. \ref{asm_module}). The unsafe set $\Omega_{3}$ of this situation is defined as,
\begin{equation}
    \Omega_{3} = \{ a\in A \vert a=A^{\uparrow} or A^{\oslash}, |x_{e}^{\sigma}(a)- x_{\text{obj}}| \leq d_x \},
\end{equation}
where $x_{e}^{\sigma}(a)$ is the predicted longitudinal position of the ego vehicle when executing the action $a$, \textit{i.e.}, $a=A^{\uparrow} or A^{\oslash}$. $x_{\text{obj}}$ is the longitudinal position of the adjacent vehicle that occupies the target lane. $d_x$ is the distance threshold. The condition
$|x_{e}^{\sigma}(a)- x_{\text{obj}}| \leq d_x$ indicates that the ego vehicle and the adjacent vehicle have few differences in the longitudinal direction, making it hard for the ego vehicle to merge. Similarly, the action $A^{\uparrow}$ or $A^{\oslash}$ would be replaced by the deceleration $A^{\downarrow}$, which slows down the ego vehicle to find another opportunity to merge, \textit{i.e.}, $a_{t}^{\odot} = A^{\downarrow}$. \par
Overall, we note that the ASM is designed to shield unexpected or unsafe actions, and it does not force the agent to merge. Moreover, as the constraint violation behavior is reduced, the sample efficiency of safe RL can be improved, and the convergence rate will increase as well. Despite this action shielding mechanism, the agent can still get the cost to enable it to learn to merge.

%%************** Theoretical Analysis ***************%%
\section{Theoretical Analysis}
\label{theoretical analysis}
This section presents a theoretical analysis on the ASM regarding the safety performance of the learned RL policy and the convergence performance of the proposed human-aligned safe RL algorithm. 
\subsection{Safety Performance}
\newtheorem{theorem}{\bf Theorem}
\begin{theorem}\label{theorem:qc_decrease}
    If the current action $a_t$ is unsafe, \textit{i.e.}, $a_t \in \Omega_{\text{unsafe}}$, and when it is replaced with a safe one $a_t^{\odot}$ using ASM, then the state action cost function $Q_c(s_t, a_t^{\odot})$ is lower than the original state action cost function $Q_c(s_t, a_t)$, which indicates improved safety effectiveness contributed by the ASM.
\end{theorem}
\begin{IEEEproof}
    We consider the Bellman equation that is denoted as
    \begin{equation}
        Q_c(s_t, a_t) = c(s_t, a_t) + \gamma V_c(s_{t+1}).
    \end{equation}
    For simplicity, let $c_{t} = c(s_t, a_t)$ and ${c}_{t}^{\odot} = c(s_t, a_{t}^{\odot})$. Then we have
    \begin{equation}
        Q_c(s_t, a_t) - Q_c(s_t, a^{\odot}_{t}) = c_t - c_t^{\odot} + V_c(s_{t+1}) - V_c(s_{t+1}^{\odot}).
    \end{equation}
    Considering the first and second items in the Bellman equation, the original decision $a_t$ will lead to a collision in the next steps as the result of $c_{t} > {c}^{\odot}_{t}$. As for the difference in state cost value, we assume that the following actions will also be replaced by a safe decision if the original is unsafe. Then, we have
    \begin{equation}
        V_c(s_{t+1}) - V_c(s_{t+1}^{\odot}) \!\!=\!\! E_{\pi} \bigl[\sum_{t=0}^{T}\gamma^{t} (c(s_{t+1}, a_{t+1}) - c(s_{t+1}^{\odot}, a_{t+1}^{\odot}))\bigr].
    \end{equation}
    During the rest steps in the episode, the original decision will be replaced by the safe one, and get a safe state. Therefore, we have
    \begin{equation}
        c(s_{t+1}, a_{t+1}) > c(s_{t+1}^{\odot}, a_{t+1}^{\odot}), t\in[0, T].
    \end{equation}
    As a consequence, we have
    \begin{equation}
        V_{c}(s_{t+1}) > V_{c}(s_{t+1}^{\odot}), t\in[0, T],
    \end{equation}
    and
    \begin{equation}
        Q_{c}(s_t, a_t) > Q_c(s_t, a^{\odot}_t), t\in[0, T].
    \end{equation}
\end{IEEEproof}

The performance of the policy with the ASM could be promoted because the safe action can reduce the state-action cost value at each step and guide the agent to choose the safe decision. Furthermore, another benefit of the proposed method is that the training process is not affected by the ASM. 

\subsection{Convergence Performance}
The optimization objective is to maximize the sum of the discounted reward while assuring the cost value satisfies the constraints. From Eq. (\ref{max-min problem}), the Lagrangian dual function is formulated as,
\begin{equation}
    d(\lambda) = \mathop{\text{min}}\limits_{\pi} -J_{R}^{\pi} + \lambda(J_{C}^{\pi} - \eta).
\end{equation}
The dual ascent is to find ${\text{max}}_{\lambda \geq 0}d(\lambda)$. During the learning process, the policy update is imperfect due to two factors. One factor is that the number of iterations is limited for computation efficiency and another factor is that the replaced decision can guarantee safety but not optimality. Hence, we assume the suboptimality of the solution $\pi^{*}$ is upper bounded as:
\begin{equation}
    -J_{R}^{\pi^{*}} + \lambda(J_{C}^{\pi^{*}} - \eta) - d(\lambda) < \epsilon.
\end{equation}
We denoted the residual of $\lambda$ before and after the imperfect update in a single step of Algorithm \ref{alg:srl with sacd} as,
\begin{equation}
    \hat{g}(\lambda, \alpha_{\lambda}) = \frac{1}{\alpha_{\lambda}}(\text{max}(0, \lambda+\alpha_{\lambda}\nabla_{\lambda}d(\lambda)) - \lambda).
\end{equation}

\newtheorem{lemma}{\bf Lemma}
\begin{lemma}
\label{lemma:lagrangian_dual_increase}
Following the imperfect dual ascent with step size $\alpha_\lambda \leq \mu$, we have:
\begin{equation}
    d(\lambda_{k+1}) \geq d(\lambda_{k}) + \frac{\alpha_{\lambda}}{2}\parallel\hat{g}(\lambda_k, \alpha_\lambda)\parallel^2 - \sqrt{\frac{2\epsilon}{\mu}} \parallel \hat{g}(\lambda_k, \alpha_\lambda)\parallel.
\end{equation}
\end{lemma}
\begin{IEEEproof}
    The proof of Lemma \ref{lemma:lagrangian_dual_increase} follows from the proof of Theorem 2.2.7 in \cite{Rockafellar_1970}. Let $\hat{g} = \hat{g}(\lambda, \alpha_{\lambda})$. We denoted the feasible set of $\lambda$ as $\mathcal{P}$. For all $\lambda' \in \mathcal{P}$, we have
    \begin{flalign*}
        & d(\lambda') - \frac{\mu}{2} \parallel \lambda' - \lambda_k \parallel ^2 \\
        & \leq d(\lambda_k) + \langle \nabla d(\lambda_k), \lambda' - \lambda_k \rangle \\
        & = d(\lambda_k) + \langle \nabla d(\lambda_k), \lambda_{k+1} - \lambda_k \rangle + \langle \nabla d(\lambda_k), \lambda' - \lambda_{k+1} \rangle \\
        & \leq d(\lambda_k) + \langle \nabla d(\lambda_k), \lambda_{k+1} - \lambda_k \rangle + \\
        & \indent \langle \hat{g}, \lambda' - \lambda_{k+1} \rangle + \sqrt{\frac{2\epsilon}{\mu}} \parallel \lambda' - \lambda_{k+1} \parallel \\
        & \leq d(\lambda_{k+1}) - \frac{\alpha_{\lambda}}{2} \parallel \hat{g}^2 \parallel + \\ 
        & \indent \langle \hat{g}, \lambda' - \lambda_{k} \rangle + \sqrt{\frac{2\epsilon}{\mu}} \parallel \lambda' - \lambda_{k+1} \parallel.
    \end{flalign*}
\end{IEEEproof}
Taking $\lambda' = \lambda_k$ and utilizing the fact that $\lambda_{k+1} - \lambda_k = \alpha_{\lambda} \hat{g}$ gives the result.

\begin{theorem}\label{theorem:convergence}
    There exist constant $\chi > 0$ such that the imperfect update converges to a dual solution $\hat{\lambda}$ that satisfies,
    \begin{equation}
        \mathop{\text{min}}\limits_{\lambda^{*} \in \mathcal{P}^{*}} \parallel \lambda^{*} - \hat{\lambda} \parallel \leq \chi \sqrt{\frac{\epsilon}{\mu}}.
    \end{equation}
\end{theorem}
\begin{IEEEproof}
    Let $\phi(\lambda) = \text{min}_{\lambda^{*} \in \mathcal{P}^{*}} \parallel \lambda - \lambda^{*} \parallel$. Based on the Theorem 4.1 of \cite{Luo1993OnTC},  there exists a constant $\psi$ such that
    \begin{equation}
        \phi(\lambda_k) + \parallel \pi^{*} - \pi \parallel \leq \psi \parallel \lambda_{k+1} - \lambda_k \parallel.
    \end{equation}
    From Lemma \ref{lemma:lagrangian_dual_increase}, when $\parallel \hat{g} \parallel > \frac{2}{\alpha_\lambda}\sqrt{\frac{2\epsilon}{\mu}}$, the Lagrangian dual function $d(\lambda)$ monotonically increases and the imperfect dual ascent would reach a $\hat{\lambda}$ satisfying $\parallel \hat{g}(\hat{\lambda}, \alpha_{\lambda}) \parallel < \frac{2}{\alpha_{\lambda}} \sqrt{\frac{2\epsilon}{\mu}}$. Then, we have $\parallel g(\hat{\lambda}, \alpha_{\lambda}) \parallel < (\frac{2}{\alpha_{\lambda}} + 1)\sqrt{\frac{2\epsilon}{\mu}}$. Then, it follows that $\psi \parallel \lambda_{k+1} - \lambda_k \parallel < \psi (2+\alpha_\lambda)\sqrt{\frac{2\epsilon}{\mu}}$. Taking $\psi = \frac{\chi}{\sqrt{2}(2+\alpha_\lambda)}$, we have $\phi(\hat{\lambda}) \leq \chi \sqrt{\frac{\epsilon}{\mu}}$.
\end{IEEEproof}
Theorem \ref{theorem:convergence} shows that the proposed method will converge to a near-optimal solution even under imperfect policy updates. \par

%% ************** Implementation Details ***********%%
\section{Implementation Details}
\label{Implementation Details}
This section introduces the implementation details of the experiments\footnote{Source codes are available at \href{https://github.com/wenqing-2021/On_Ramp_Merge_Safe_RL}{https://github.com/wenqing-2021/On\_Ramp\_Merge\_Safe\_RL}}, including the simulation environment, scenario setting, model formulation, and implementation details, \textit{etc}.

%% Table: Hyperparameters
\begin{table}[b]
\caption{Hyperparameters of RL \label{table:hyperparameters}}
\renewcommand\arraystretch{1.25}
\centering
\begin{tabular}{ l c }
\bottomrule[1pt]
\makebox[0.23\textwidth][l]{Parameter} & \makebox[0.1\textwidth][c]{Value}\\
\bottomrule[1pt]
Optimizer & Adam \\
Policy network learning rate $\alpha_{\theta}$ & 1e-4 \\

Critic network learning rate $\alpha_{\omega}$ & 1e-4 \\

Cost network learning rate $\alpha_{\omega_c}$ & 1e-4 \\

Temperature parameter learning rate $\alpha_{\xi}$& 1e-4 \\

Initial Lagrangian multiplier $\lambda_{0}$ & 1.0 \\

Lagrangian multiplier learning rate $\alpha_{\lambda}$ & 1e-4 \\

Replay Buffer size $\mathcal{D}$ & 1e5 \\

Batch size $\mathcal{B}$& 256 \\
\bottomrule[1pt]
\end{tabular}
\end{table}

%% figure: Training performances of our method and the baseline methods
\begin{figure*}[t]
\centering
\subfloat[Comparative study on Dueling DQN, SACD, PPO, and SACD-$\lambda$-TM (ours)]{\includegraphics[width=6in,trim=0 5 0 0, clip]{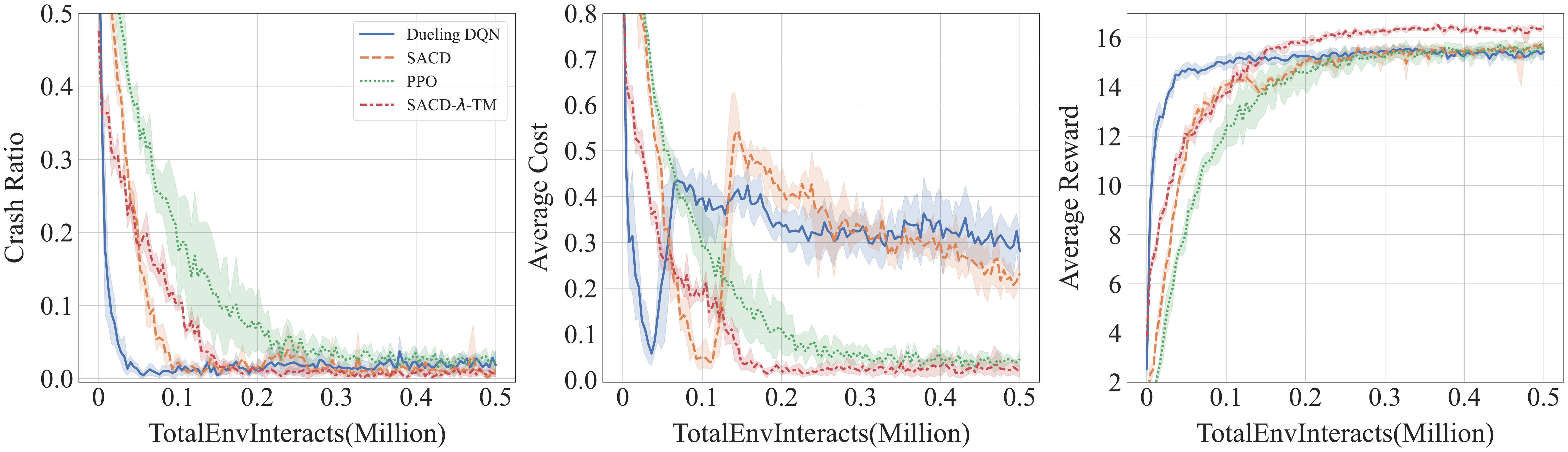}}
\vspace{-1mm}
\subfloat[Ablation study on SACD, SACD-$\lambda$, SACD-$\lambda$-M, and SACD-$\lambda$-TM (ours)]{\includegraphics[width=6in,trim=0 5 0 0, clip]{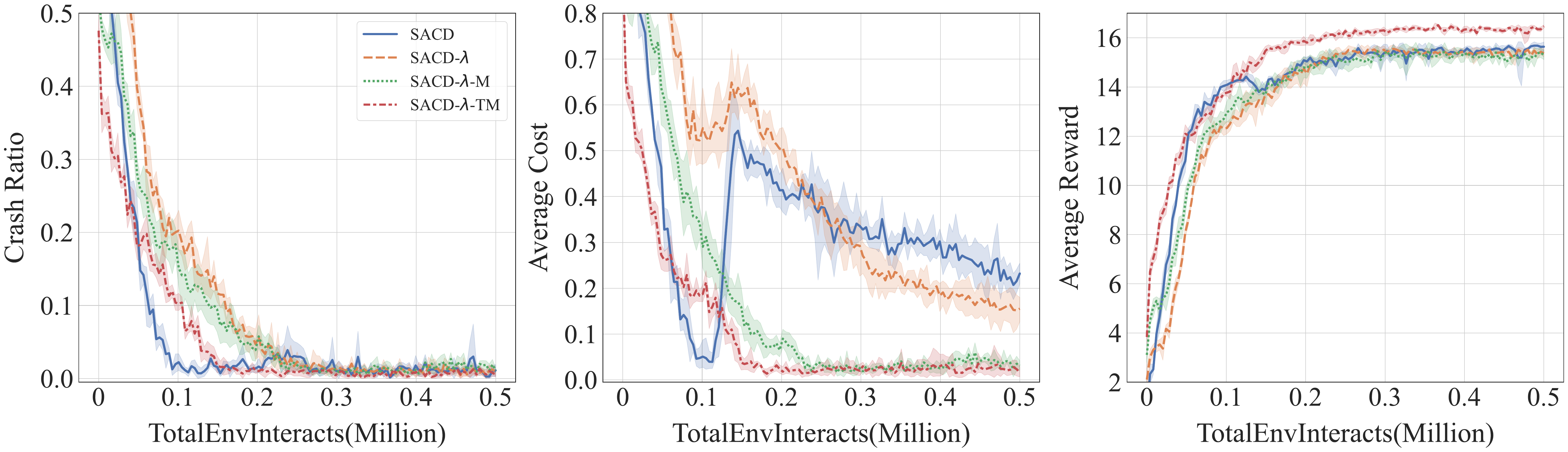}}
\caption{Training process of our method and the baseline methods in (a) comparative study, (b) ablation study. 
The solid line is the average value and the shadow area is the confidence interval of 95$\%$ over five runs of different random seeds.}
\label{experimental results}
\end{figure*}

%% table: comparative with RL methods
\begin{table*}[t]
\caption{Comparison between the constraint-free RL methods and ours.
\label{table:rl}}
\renewcommand\arraystretch{1.8}
\centering
% \resizebox{6in}{3.5cm}{
\setlength{\tabcolsep}{2.5mm}{
\begin{tabular}{c c c c c c c c c c c c c}
\bottomrule[1pt]
\multirow{2}{*}{Method} & \multicolumn{3}{c}{Success Rate ($\%$) } & \multicolumn{3}{c}{Collision Rate} & \multicolumn{3}{c}{Average Cost} & \multicolumn{3}{c}{Average Time (s)}  \\
&\multicolumn{1}{c}{High} & Medium & Low & \multicolumn{1}{c}{High} & Medium & Low & \multicolumn{1}{c}{High} & Medium & Low & \multicolumn{1}{c}{High} & Medium & Low \\
\bottomrule[1pt]
Dueling DQN & 87.0 & 94.3 & 99.0
& 0.013 & 0.005 & 0.005
& 0.50 & 0.28 & 0.08 
& 11.78 & 11.47 & 10.85 \\
\hline
SACD & 94.5 & 97.5  & 99.2 
& 0.010 & 0.008 & 0.005
& 0.44 & 0.25 & 0.10
& 11.59 & 11.33 & 10.82 \\
\hline
PPO & 99.5 & 97.7 & 99.2
& 0.003 & 0.018 & 0.008
& 0.01 & 0.03 & 0.01
& 12.36 & 11.62 & 10.95 \\
\hline
\textbf{SACD-\boldsymbol{$\lambda$}-TM (ours)} & \textbf{99.0} & \textbf{99.5} & \textbf{99.3}
& \textbf{0.003} & \textbf{0.005} & \textbf{0.005}
& \textbf{0.02} & \textbf{0.02} & \textbf{0.02}
& \textbf{11.87} & \textbf{11.46} & \textbf{10.96}\\
\bottomrule[1pt]
\end{tabular}}
\end{table*}

%% table: Lattice and ours
\begin{table*}[t]\tiny
\caption{Comparison between lattice-based method and ours.
\label{table:lattice and ours}}
\renewcommand\arraystretch{1.8}
\centering
\resizebox{2\columnwidth}{!}{
\begin{tabular}{c c c c c c c c c c c c c}
\bottomrule[0.5pt]
\multirow{2}{*}{Method} & \multicolumn{3}{c}{Success Rate ($\%$) } & \multicolumn{3}{c}{Collision Rate} & \multicolumn{3}{c}{Average Time (s)}  \\
&\multicolumn{1}{c}{High} & Medium & Low & 
\multicolumn{1}{c}{High} & Medium & Low &  
\multicolumn{1}{c}{High} & Medium & Low \\
\bottomrule[0.15pt]
\hline
Lattice & 98.5 & 99.0 & 97.3
& 0.012 & 0.010 & 0.027
& 10.44 & 10.12 & 9.93 \\
\hline
\textbf{SACD-\boldsymbol{$\lambda$}-TM (ours)} & \textbf{99.0} & \textbf{99.5} & \textbf{99.3}
& \textbf{0.003} & \textbf{0.005} & \textbf{0.005}
& \textbf{11.87} & \textbf{11.46} & \textbf{10.96}\\
\bottomrule[0.5pt]
\end{tabular}}
\end{table*}

%% table: ablations study
\begin{table*}[t]
\caption{Ablation study.
\label{table:ablations}}
\renewcommand\arraystretch{1.8}
\centering
% TABLE V
\resizebox{2\columnwidth}{!}{
\begin{tabular}{c c c c c c c c c c c c c}
\bottomrule[1pt]
\multirow{2}{*}{Method} & \multicolumn{3}{c}{Success Rate ($\%$) } & \multicolumn{3}{c}{Collision Rate} & \multicolumn{3}{c}{Average Cost} & \multicolumn{3}{c}{Average Time (s)}  \\
&\multicolumn{1}{c}{High} & Medium & Low & \multicolumn{1}{c}{High} & Medium & Low & \multicolumn{1}{c}{High} & Medium & Low & \multicolumn{1}{c}{High} & Medium & Low \\
\bottomrule[1pt]
SACD & 94.5 & 97.5  & 99.2 
& 0.010 & 0.008 & 0.005
& 0.44 & 0.25 & 0.10
& 11.59 & 11.33 & 10.82 \\
\hline
SACD-$\lambda$ & 97.2 & 97.3 & 99.3
& 0.003 & 0.005 & 0.005
& 0.23 & 0.13 & 0.04
& 12.08 & 11.54 & 10.99\\
\hline
SACD-$\lambda$-M & 98.3 & 98.8 & 98.9
& 0.010 & 0.008 & 0.003
& 0.02 & 0.02 & 0.02
& 12.33 & 11.58 & 11.02\\
\hline
\textbf{SACD-\boldsymbol{$\lambda$}-TM (ours)} & \textbf{99.0} & \textbf{99.5} & \textbf{99.3}
& \textbf{0.003} & \textbf{0.005} & \textbf{0.005}
& \textbf{0.02} & \textbf{0.02} & \textbf{0.02}
& \textbf{11.87} & \textbf{11.46} & \textbf{10.96}\\
\bottomrule[1pt]
\end{tabular}}
\end{table*}

\subsection{Scenario Settings} 
\label{subsubsec:scenario settings}
We built an on-ramp merging scenario based on the simulation platform \textit{highway-env} \cite{highway-env}.  As shown in Fig. \ref{fig:merge scenario}, the merge zone is comprised of a lane width of 5 meters and a length of 70 meters. The ego vehicle starts from the entrance of the ramp,  80 meters from the merge zone. The longitudinal acceleration of the surrounding vehicles is predicted by the intelligent driver model (IDM) \cite{treiber2000congested}. The longitudinal distance  between the front vehicle $i$ and the rear vehicle $i+1$ is denoted as $d_{i,i+1}$, which is defined as,
\begin{equation}
\label{exprimental disctance design}
d_{i,i+1} = d_s + \rho v_{i+1},
\end{equation}
where $d_s$ is safety distance. $v_{i+1}$ is the speed of the rear vehicle. The initial speed of all vehicles ranges from 17 m/s to 27 m/s. $\rho \in [0.5,1]$ represents the space density of traffic flow. The traffic density is in low level when $\rho \in [0.5,0.7)$; medium level when $\rho \in [0.5,1]$,  high level when $\rho \in (0.8,1]$.

\subsection{Implementation Details}
The parameter settings of the proposed approach are introduced, including MPC and safe RL. The prediction horizon of MPC is set as 10 time steps. As discussed previously, the steering angle and acceleration are the control variables of the low-level controller. To ensure a smooth merge, the reference value of the steering angle is 0 rad, and the max value of the steering angle is $\pi$/8 rad. The acceleration is limited to [-0.5g,0.5g]. In particular, the prediction coefficient $\sigma$ is 5, indicating that we look five steps ahead for collision check in the ASM. For safe RL, the simulation frequency of the environment is 10Hz, and the decision-making frequency is 2Hz. Throughout the training phase, the episode terminates either upon the ego vehicle reaching the goal point or a collision occurs. The agent interacts with the environment for 0.5 million steps to gather data regarding the state, action, reward, and cost signals. We note that the agent is coupled with MPC online during the training process. Specifically, the agent receives the observation and generates an action for MPC to make motion predictions. If the predicted position indicates a collision, the ASM replaces the original action with a safe one. Finally, the agent executes the safe action in the low-level MPC controller, garnering both reward and cost from the environment. All neural networks used in safe RL have 2 hidden layers with 256 units per layer, and the activation function is ReLU. The optimizer and learning rate of the network parameters used in the RL are summarized in Table \ref{table:hyperparameters}. Moreover, we train the RL agent with five random seeds. All the experiments are conducted on an Intel Core i5-11300H CPU with 16 GB RAM that runs at 3.1 GHz.

\subsection{Baselines}
In this study, the proposed human-aligned safe RL algorithm is denoted as SACD-$\lambda$-TM. To validate the effectiveness of our method, the proposed approach is compared with the following baseline methods.
\begin{itemize}
    \item \textbf{Lattice} \cite{werling2010optimal}: a sample-based motion planning algorithm, which simplifies the planning problem by discretizing the continuous state space into a finite set of discrete states, making it easier to search for the optimal path from an initial state to a goal state.
    \item \textbf{Dueling DQN} \cite{wang2016dueling}: one variant of Deep Q Network (DQN), which has two separate estimators, one for the state value function and one for the state-dependent action advantage function. This architecture leads to better policy evaluation in the presence of many similar-valued actions.
    \item \textbf{PPO} \cite{SchulmanWDRK17}: it uses a surrogate objective function that combines the advantages of the current policy and the ratio of the probability of the new policy to the old policy.
    % PPO introduces a clipping mechanism to limit the policy change, which helps avoid the policy updates that could result in divergent behavior.
    Proximal Policy Optimization (PPO) introduces a clipping mechanism to limit the policy change, which helps avoid the divergent behavior.
\end{itemize}

In addition, we have made an ablation study based on the following algorithms, including a) SACD, a discrete action version of a SAC algorithm without the constrained settings; b) SACD-\boldsymbol{$\lambda$} is a modified version of SACD, which utilizes the Lagrange method to solve the CMDP problem, without the ASM; c) SACD-\boldsymbol{$\lambda$}-M, which has the ASM that replaces unsafe actions with safe ones; d) SACD-\boldsymbol{$\lambda$}-TM, which uses n-step TD prediction for estimating the soft Q-value functions.

\subsection{Evaluation Metrics}
The proposed approach is evaluated in terms of safety and traffic efficiency. Specifically, safety metrics focus on the average cost and the number of collisions. The traffic efficiency metrics include a) the average reward, which presents the traffic efficiency. The higher the average reward, the higher the efficiency of the ego vehicle; b) the success ratio, which is the percentage of safely reaching the goal with accumulated cost less than 0.5; c) average time, which is the time takes to merge into the target lane; d) average cost, which presents the driving risk level. The higher the average cost, the more dangerous the driving situation would be.

%% ************** Results and Discussion ***********%%
\section{Results and Discussion}
\label{results}
In this section, we evaluate our method regarding the learning efficiency, safety, and traffic efficiency of the decision policy. We also conducted an ablation study to identify the essential components of our approach. 

% \subsubsection{Real-time Peformance}
% The proposed method is evaluated in different traffic densities and the computational time of each decision-making step is less than 0.05s satisfying the real-time performance, as shown in Fig \ref{fig:decision_time}. 

\subsection{Comparative Study}
In this section, we compare the proposed method with several popular RL algorithms, and we also evaluate the learned decision policy with the traditional sample-based planners and human-driven vehicles. 
\subsubsection{\textbf{SACD-$\lambda$-TM versus. RL-based methods}} The proposed approach SACD-$\lambda$-TM is compared with several popular RL-based methods, such as Dueling DQN, SACD, and PPO, concerning the success rate, collision rate, average cost, and average time to merge. As shown in Fig. \ref{experimental results} (a), our method SACD-$\lambda$-TM outperforms all baselines and achieves the highest average reward but the lowest average costs. Dueling DQN has the fastest convergence rate but has the largest cost, indicating the highest risks. Compared with PPO, our method converges faster and achieves a higher average reward. Similarly, SACD has poor performance in terms of average cost. Overall, the proposed SACD-$\lambda$-TM outperforms the baseline methods regarding convergence rate and safety level. Also, we conduct quantitative evaluations regarding success ratio, collision rate, average cost, and average time, as shown in Table \ref{table:rl}. {We trained the agents in the medium density of traffic and then evaluated them over 400 episodes in the high, medium, and low density of traffic, respectively. From Table \ref{table:rl}, our method has achieved the minimum collision rate among all baselines. The dueling DQN and SACD have the highest collision rate in the high density of traffic. PPO and our method SACD-$\lambda$-TM have seen the highest collision rate in the medium density of traffic. The collision rates of our method in high, medium, and low-density traffic are 0.003, 0.005, and 0.005, and the average number of collisions in the high, medium, and low-density traffic are 1.2, 2, and 2, which has not shown a big difference. Therefore, we can see that the collision rate is not sensitive to the traffic density in our approach, and the average cost also does not change with the traffic density. In contrast, the average time and success rate in low-density traffic outperformed those in high- and medium-density traffic, showing that the complexity of the traffic density mainly impacts the success rate and traffic efficiency in our approach. Our method has achieved the highest success rate (99.5\%) and the lowest cost (0.02) in the medium density of traffic. Fig. \ref{fig:cost_distribution} illustrates the cost distribution of the proposed methods and baseline algorithms. We observe that more than 40\% of the cost of Dueling DQN and SACD is higher than 0.5, while nearly 95\% of the cost of PPO and SACD-$\lambda$-TM is lower than 0.5, showing improved safety. Despite the fact that PPO has achieved comparable performance with our method in terms of safety cost, PPO converges slower and yields less traffic efficiency than our method.}

\subsubsection{\textbf{SACD-$\lambda$-TM versus. sample-based methods}} 
we compare the proposed method SACD-$\lambda$-TM with a sample-based method, \textit{i.e.}, Lattice, and the quantitative results are shown in Table \ref{table:lattice and ours}. From the comparison, our method can achieve a higher success rate and lower collision rate than the sample-based method in different densities of traffic. However, it takes less time for the Lattice-based method to merge. Moreover, the Lattice-based method is computationally demanding in complex scenarios, as a large number of candidate trajectories need to be generated to ensure the feasibility of the solution, which may not meet the real-time requirements of the system.

\subsubsection{\textbf{SACD-$\lambda$-TM versus. human-driven vehicle}} 
\begin{figure}[t]
\centering
\includegraphics[width=3.5in,trim=0 0 0 0, clip]{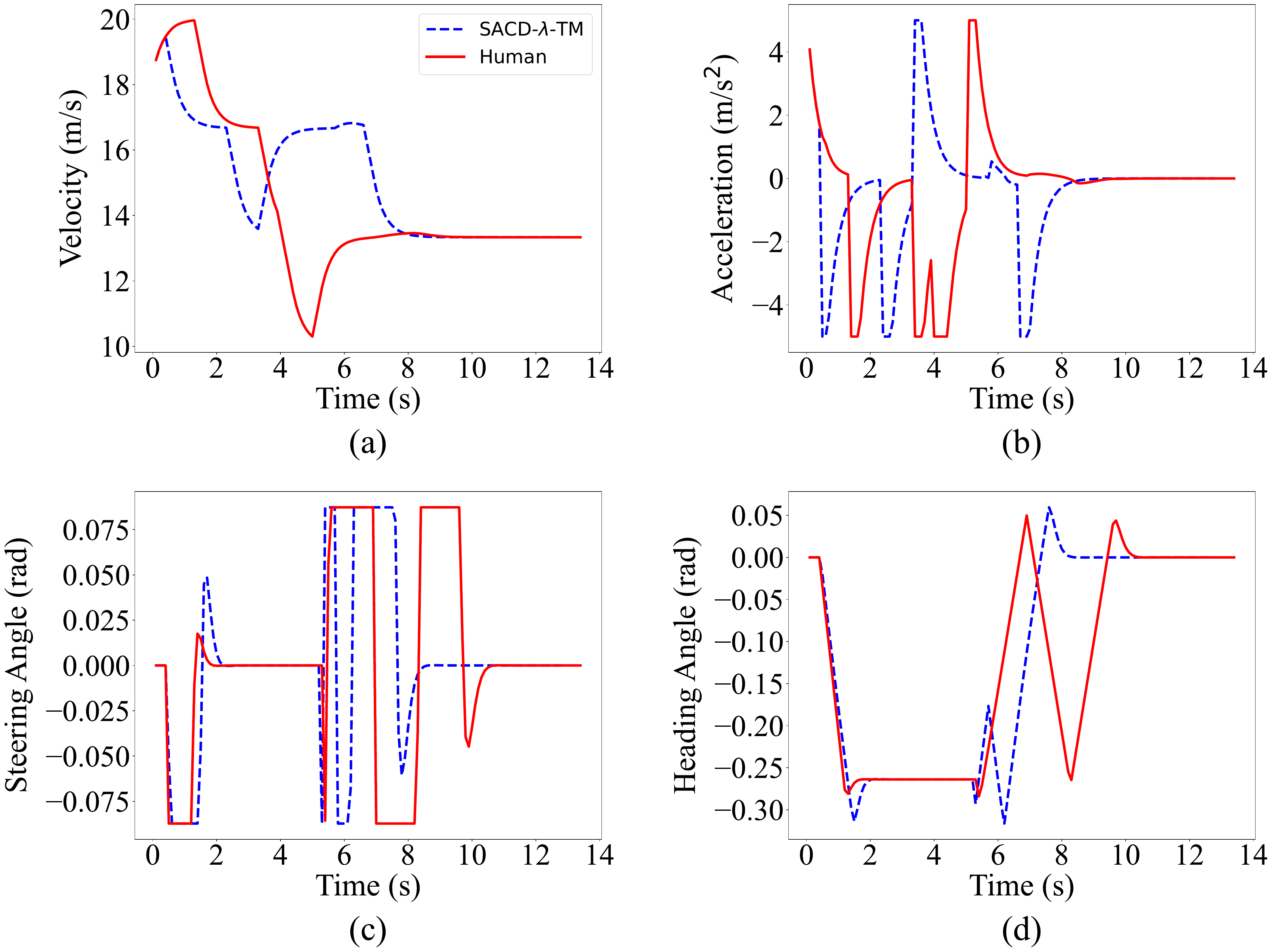}
\caption{The kinematic parameters vary with time after merging. We test the ego vehicle in the medium traffic density with the same scenario settings to compare the proposed method with the human-driven vehicle.}
\label{fig:human_rl_compare}
\end{figure}

we compare the proposed method with the human-driven vehicle, \textit{i.e.}, using the directional arrow keys on the keyboard to control the ego vehicle. The velocity, acceleration, steering angle, and yaw angle of the ego vehicle are presented in Fig. \ref{fig:human_rl_compare}. As shown in Fig. \ref{fig:human_rl_compare} (a), the proposed method has a higher average speed compared with the human-driven vehicle, and there is no obvious difference in the acceleration curves (see Fig. \ref{fig:human_rl_compare} (b)). Besides, Fig. \ref{fig:human_rl_compare} (c) and (d) are the steering angle and yaw angle of the ego vehicle, showing that our method can enable a faster merge than the human-driven vehicle. \par

{Overall, compared to other RL methods (i.e., Dueling DQN, SACD, and PPO), our approach demonstrates advantages in collision rate, average cost, average reward, and convergence speed, as shown in Fig. \ref{experimental results} and Table \ref{table:rl}. Besides, our approach shows higher success rates and lower collision rates than the sample-based method (i.e., Lattice), as shown in Table \ref{table:lattice and ours}. Additionally, compared to the human-driven method, our approach achieves faster onto-ramp merging (see Fig. \ref{fig:human_rl_compare}).} 

{
Although we and \cite{chen2023} share a similar architecture for on-ramp merging, \textit{i.e.}, RL for the high-level decision and MPC for a low-level controller, there exist many differences that make it hard to compare the two approaches fairly and determine which one is better based on the collision rate. First, the two studies are trained in different environments for different tasks. Our method focuses on the single-agent RL, while \cite{chen2023} is intended for multi-agent learning, and it aims to train the multiple agents simultaneously to cooperate to enable autonomous merging. In other words, \cite{chen2023} is designed for the on-ramp merging with multiple connected and automated vehicles, and our method is intended for the single AV. Besides, the length of the merge zone is shorter in this study, indicating that our simulation environment exhibits higher complexity and more challenges, making it hard for us to achieve zero collision. Moreover, we notice that zero collision can only be realized with proper hyper-parameter tuning in \cite{chen2023}, which is hard to generalize to new environments as well. In addition, we use CDMP and a shielding mechanism, which can help significantly reduce safety violations during the training process and is more suitable for safety-critical applications. We also provide theoretical proof to validate the shielding mechanism's effectiveness in enhancing RL's safety and sample efficiency. Therefore, comparing our method with \cite{chen2023} regarding collision rate is not applicable. Similarly, our collision rate is very low (\textit{e.g.}, collision rate of 0.05 in the medium traffic mode), which indicates 2 collisions for 400 runs. Also, we observe that our method always yields smaller collision rates than the baseline methods, demonstrating the effectiveness of our approach. } 

{To further reduce the collision rate, we try to provide several solutions that might be useful in future research. First, we notice that \cite{chen2023} achieves zero collision when the hyper-parameters are well-tuned. Inspired by that, we believe an automatic optimization of the hyper-parameter can be made to find proper hyper-parameter settings for the cost terms of safety, the threshold of the cost limits, and the prediction steps of the ASM, which can help reduce the collision rate. Besides, using conditional value at risk or other risk measures to optimize policies that account for the tail risk of collisions would also be helpful, focusing on minimizing the worst-case outcomes \cite{WCSAC}. Incorporating the collision constraints into the low-level MPC can also reduce collisions in the trajectory planning stage, while the computational complexity might increase as well. }
%% cost distribution
\begin{figure}[t]
    \centering
    \includegraphics[width=2.2in, trim= 0 0 0 0, clip]{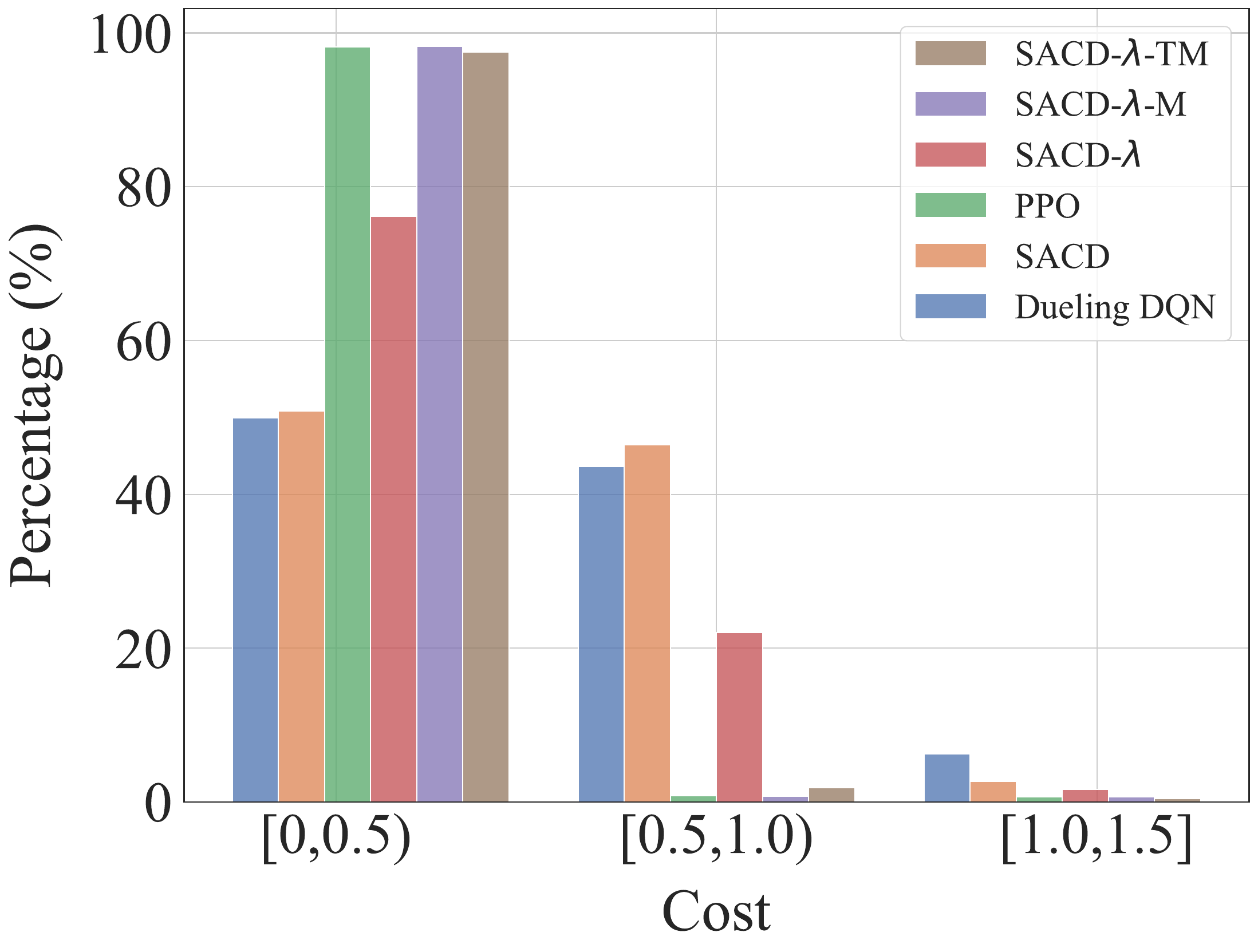}
    \caption{The histogram of the average cost of the proposed method SACD-$\lambda$-TM and all baseline algorithms.}
    \label{fig:cost_distribution}
\end{figure}
%% human compare
%% visualization
\begin{figure*}[t]
\centering
\subfloat[SACD-$\lambda$]{\includegraphics[width=6.5in,trim=0 300 0 350, clip]{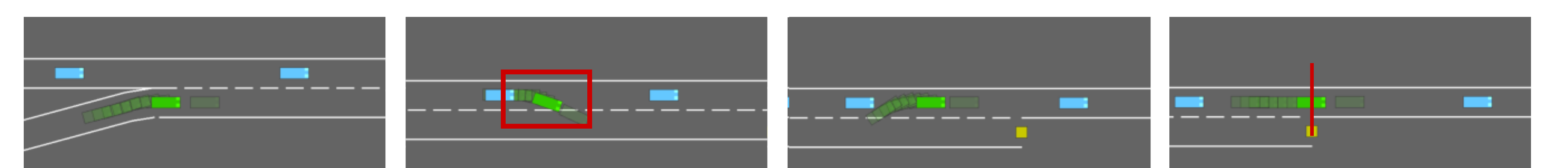}
\label{animation sacd-baseline}}
\vspace{-3mm}
\subfloat[SACD-$\lambda$-M]{\includegraphics[width=6.5in,trim=0 300 0 350, clip]{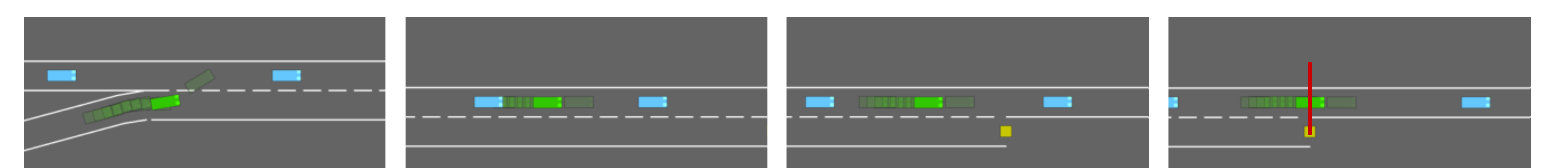}
\label{animation sacd-mpc}}
\vspace{-3mm}
\subfloat[SACD-$\lambda$-TM]{\includegraphics[width=6.5in,trim=0 230 0 350, clip]{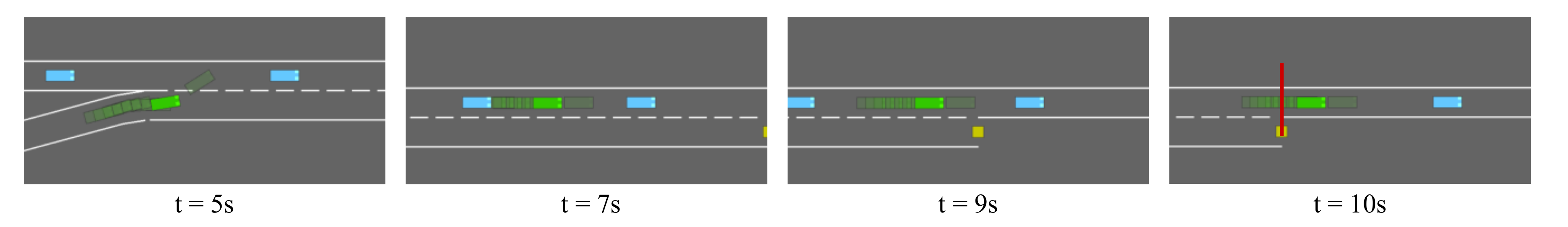}
\label{animation sacd-mpc-nsteps}}
\caption{Visualization of the trained agent, including (a) SACD-$\lambda$, (b) SACD-$\lambda$-M, and (c) SACD-$\lambda$-TM. The ego vehicle (green) trained by SACD-$\lambda$ turns right even if has merged (t=7s). In particular, our method SACD$\lambda$-TM reaches the farthest position at t=10s, indicating improved traffic efficiency}
\label{animaition process}
\end{figure*}
% %% decision time (暂不考虑实时性)
% \begin{figure}[t]
% \centering
% \includegraphics[width=2.5in,trim=0 0 0 0, clip]{pictures/section_6/Decision_Time.pdf}
% \caption{The computational time of each decision-making step within one episode in different traffic densities. All of the decision-making processes are conducted within 0.05 seconds guaranteeing real-time performance.}
% \label{fig:decision_time}
% \end{figure}

%%
\subsection{Ablation Study}
This section evaluates the impact of essential components of the proposed method SACD-$\lambda$-TM, \textit{i.e.}, the $n$-step TD prediction, ASM, and safety constraints. We compare SACD-$\lambda$-TM with three ablations, including SACD (w/o safety constraints), SACD-$\lambda$ (w/o ASM), and SACD-$\lambda$-M (w/o $n$-step TD prediction), as shown in Table \ref{table:ablations}. From Fig. \ref{experimental results} (b), our method SACD-$\lambda$-TM achieves the highest rewards, fewest costs, and smallest crash ratio among all. We also evaluate the effects of MPC in the ASM. The trained agents of SACD-$\lambda$, SACD-$\lambda$-M, and SACD-$\lambda$-TM are visualized for comparison as well.  
\subsubsection{\textbf{n-step TD Prediction}}
SACD-$\lambda$-TM utilizes $n$-step TD prediction to reduce the value estimation error and enhance the training process, while SACD-$\lambda$-M employs only one-step TD prediction. From Fig. \ref{experimental results} (b), SACD-$\lambda$-TM converges faster and achieves a higher reward than SACD-$\lambda$-M, indicating the improved learning efficiency and traffic efficiency due to the use of the $n$-step TD prediction. Furthermore, Table \ref{table:ablations} shows that SACD-$\lambda$-TM has a higher success rate, lower collision rate, and less time to merge than the baseline SACD-$\lambda$-M, also indicating the effectiveness of $n$-step TD prediction in improving safety and efficiency.
\subsubsection{\textbf{ASM}} 
we use the ASM to prevent unsafe or unexpected actions that could lead to task failure or collisions. The baseline SACD-$\lambda$ is compared with the baseline SACD-$\lambda$-M and our method SACD-$\lambda$-TM to identify the effects of the ASM. From Fig. \ref{experimental results} (b) and Table \ref{table:ablations}, SACD-$\lambda$ and SACD-$\lambda$-TM have similar values of the crash ratio and collision rate when they converge. However, SACD-$\lambda$-TM has higher rewards and lower costs than SACD-$\lambda$. Table \ref{table:ablations} also shows a higher success rate and lower cost of SACD-$\lambda$-TM than SACD-$\lambda$. Moreover, the comparison between SACD-$\lambda$-M and SACD-$\lambda$ regarding the average cost also indicates that the safety performance has been enhanced by the ASM.
\subsubsection{\textbf{Safety Constraints}}
SACD-$\lambda$-TM employs the Lagrange multiplier $\lambda$ to address the constrained RL problem. To investigate the impact of safety constraints, we compare SACD-$\lambda$ with SACD. The training progress of SACD and SACD-$\lambda$ is depicted in Fig. \ref{experimental results} (b), showing that SACD converges faster than SACD-$\lambda$, while SACD-$\lambda$ has a lower cost than SACD. The cost distribution of SACD-$\lambda$ in Fig. \ref{fig:cost_distribution} indicates a significant reduction in costs between 0.5 and 1.0 compared to SACD. Moreover, Table \ref{table:ablations} shows that SACD-$\lambda$ outperforms SACD in terms of success rate, collision rate, and cost reduction in high-density traffic scenarios. Overall, the substantial decrease in average cost demonstrates the safety advantages of incorporating safety constraints. 

We visualize the agents trained by SACD-$\lambda$, SACD-$\lambda$-M, and SACD-$\lambda$-TM for the on-ramp merging task, as illustrated in Fig. \ref{animaition process}. The ego vehicle learned by SACD-$\lambda$ may take unexpected action, \textit{e.g.}, turning right even though it has merged with the main lane (see Fig. \ref{animaition process} (a)). With the ASM, the unexpected action in SACD-$\lambda$-M and SACD-$\lambda$-TM can be avoided (see Fig. \ref{animaition process} (b) and (c)). In addition, the agent trained by SACD-$\lambda$-TM reaches the farthest position from the end of the merge zone when the episode ends, indicating improved traffic efficiency benefited from the $n$-step TD prediction.

\subsubsection{\textbf{Motion Prediction Module}}
\begin{figure}[t]
\centering
\includegraphics[width=3.5in,trim=0 0 0 0, clip]{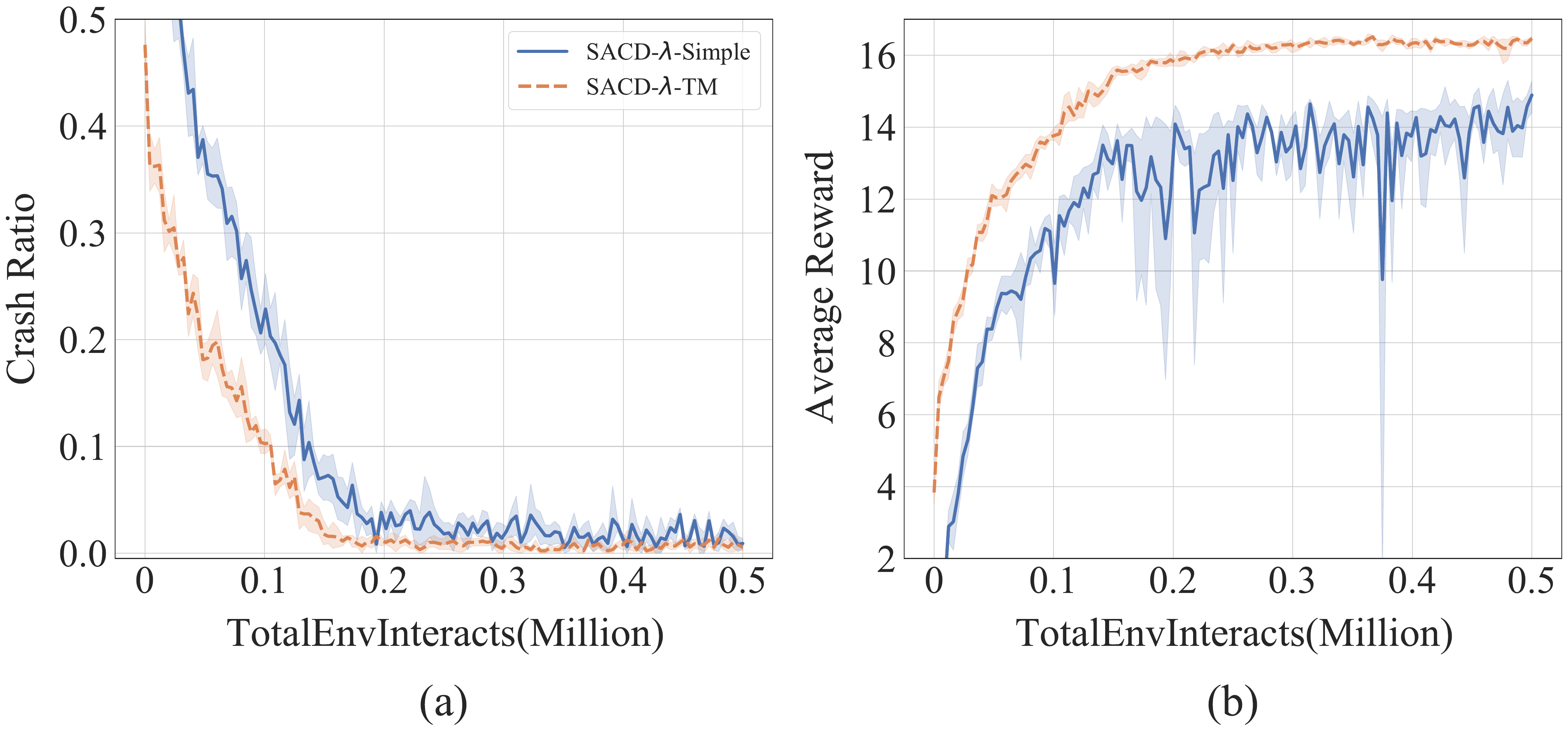}
\caption{The training result of different motion prediction methods. The proposed method based on MPC has a higher average reward, indicating its traffic efficiency is greater than that of another method using the simple prediction method for training.}
\label{fig:motion_predictive_ablation_study}
\end{figure}

we use MPC to predict the future states of the ego vehicle for collision check in the ASM of SACD-$\lambda$-TM. To investigate the effects of the motion prediction module, we use a simple kinematics-based motion prediction model for comparison, \textit{i.e.}, SACD-$\lambda$-Simple, as shown in Fig. \ref{fig:motion_predictive_ablation_study}. The proposed method SACD-$\lambda$-TM converges faster and yields a higher reward than SACD-$\lambda$-Simple, indicating improved learning efficiency and safety.

%% ablation study: mpc

\subsection{Sensitivity Analysis}
To evaluate the sensitivity of our method concerning important hyperparameters, we analyze the effects of two specific parameters, \textit{i.e.}, the prediction coefficient $\sigma$ in the ASM, and the cost limit $\eta$ in CMDP.
\subsubsection{\textbf{Prediction Coefficient}}
\label{prediction_discussion}
\begin{figure}[t]
\centering
\includegraphics[width=3.5in,trim=0 0 0 0, clip]{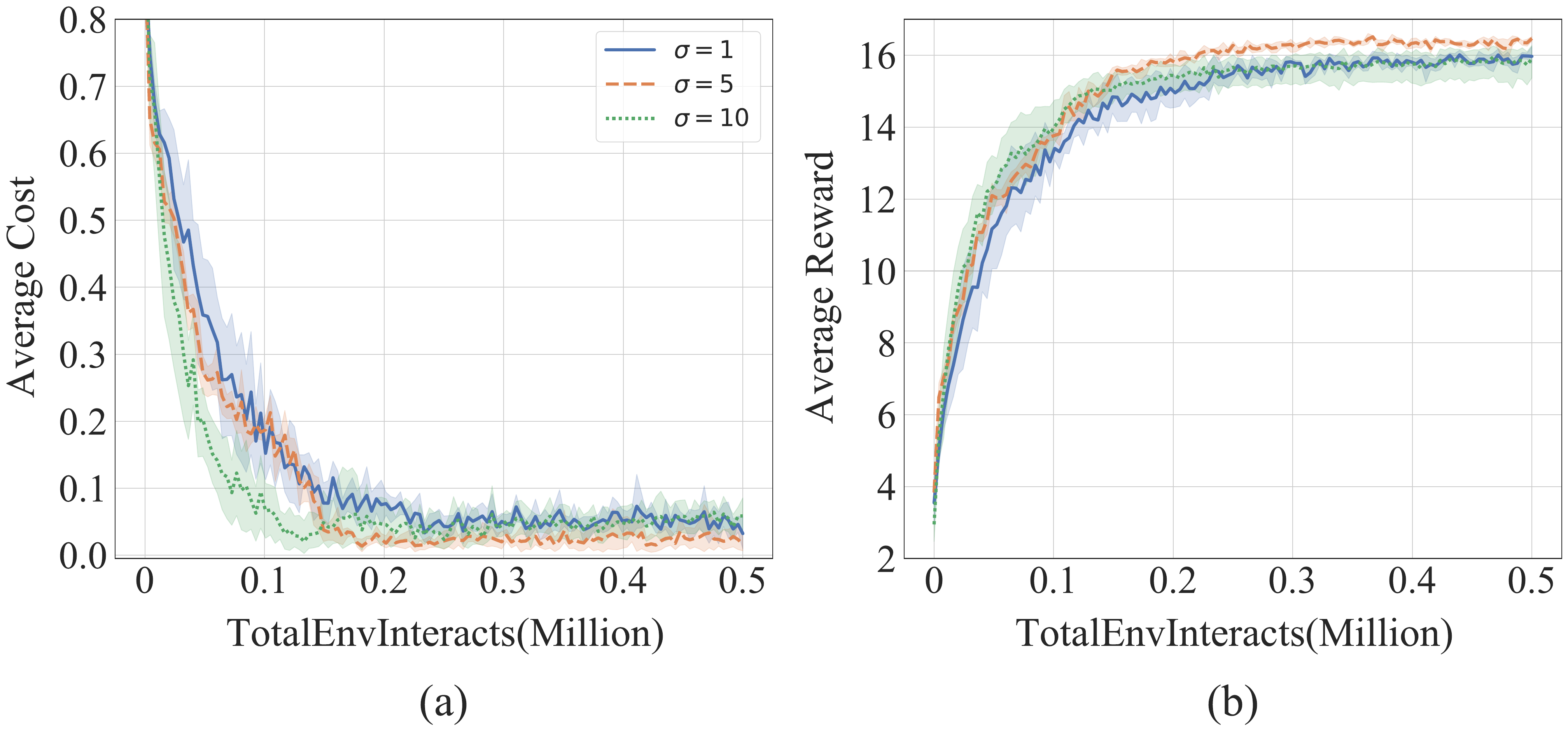}
\caption{Sensitivity of SACD-$\lambda$-TM to the prediction coefficient, \textit{i.e.}, $\sigma=1,5,10$. The solid line is the average value, and the shadow area is the confidence interval of 95 $\%$ over five runs with different random seeds.}
\label{sigma experimental results}
\end{figure}

we conduct motion prediction for constraint checking in the ASM of SACD-$\lambda$-TM and change the prediction coefficient $\sigma$ to test its effect on the average cost and average reward. As shown in Fig. \ref{sigma experimental results}, we conduct five runs for each $\sigma$, initialized with different random seeds. We observe that a larger $\sigma$ leads to faster learning but results in higher variability across the random seeds, indicating increased instability. On the other hand, a smaller $\sigma$ can slow down the training process. However, with an appropriate prediction coefficient of $\sigma$=5, the model can learn faster and achieve the best performance regarding the average cost and the average reward.
 
% sigma experimental results

\subsubsection{\textbf{Cost Limit}}
\begin{figure}[t]
\centering
\includegraphics[width=3.5in]{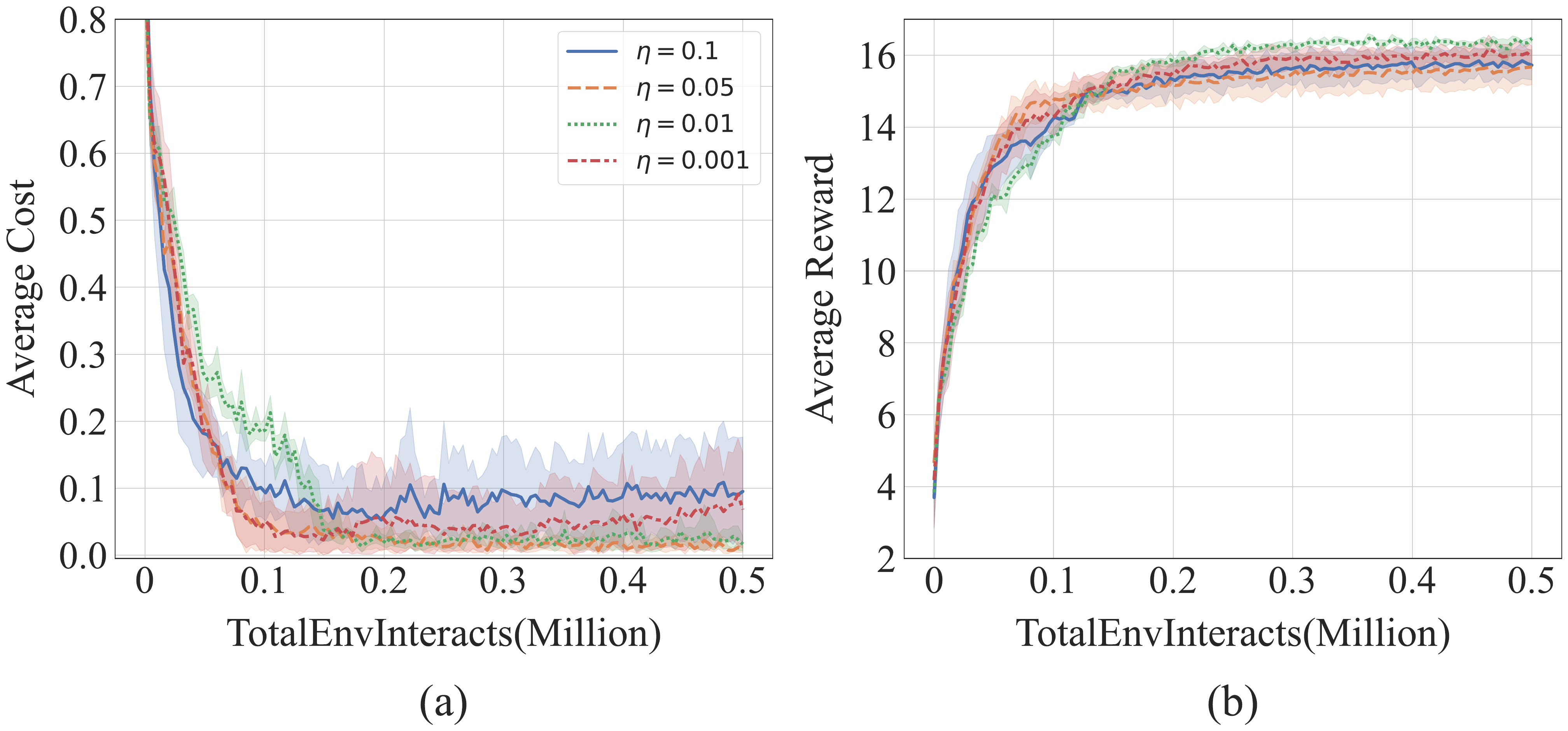}
\caption{Sensitivity of SACD-$\lambda$-TM to the hyperparameter cost limit $\eta$, where $\eta=0.1, 0,05, 0.01, 0.001$. (a) Large $\eta$ can lead to instabilities and substantial increases in the average cost, while small $\eta$ can make training slower. (b) The average reward changes slightly when $\eta$ varies. Both the average cost and the average reward can achieve the best performance when $\eta=0.01$.}
\label{cost limit influence}
\end{figure}

we use CMDP to formulate the high-level decision-making problem, and the parameter cost limit $\eta$ determines the maximum allowed accumulative cost in the constraints. To investigate the influence of the cost limit on the learning process and performance, we set the cost limit $\eta$ as 0.1, 0.05, 0.01, and 0.001, as shown in Fig. \ref{cost limit influence}. Simulation results indicate that the average cost is highly sensitive to the value of $\eta$, as it is the threshold for constraining the average cost during training. Large $\eta$ can lead to instabilities and significant increases in the average cost, while small $\eta$ can make training slower. However, there exist slight changes in the average reward when $\eta$ varies, suggesting that the cost limit can be tuned independently to achieve the best safety performance without affecting the rewards. Moreover, we observe that both the average cost and the average reward can achieve the best performance when $\eta=0.01$. 

% cost limit

\subsection{Generalization Ability}
\begin{figure}[t]
    \centering
    \includegraphics[width=3.5in, trim=0 0 0 0, clip]{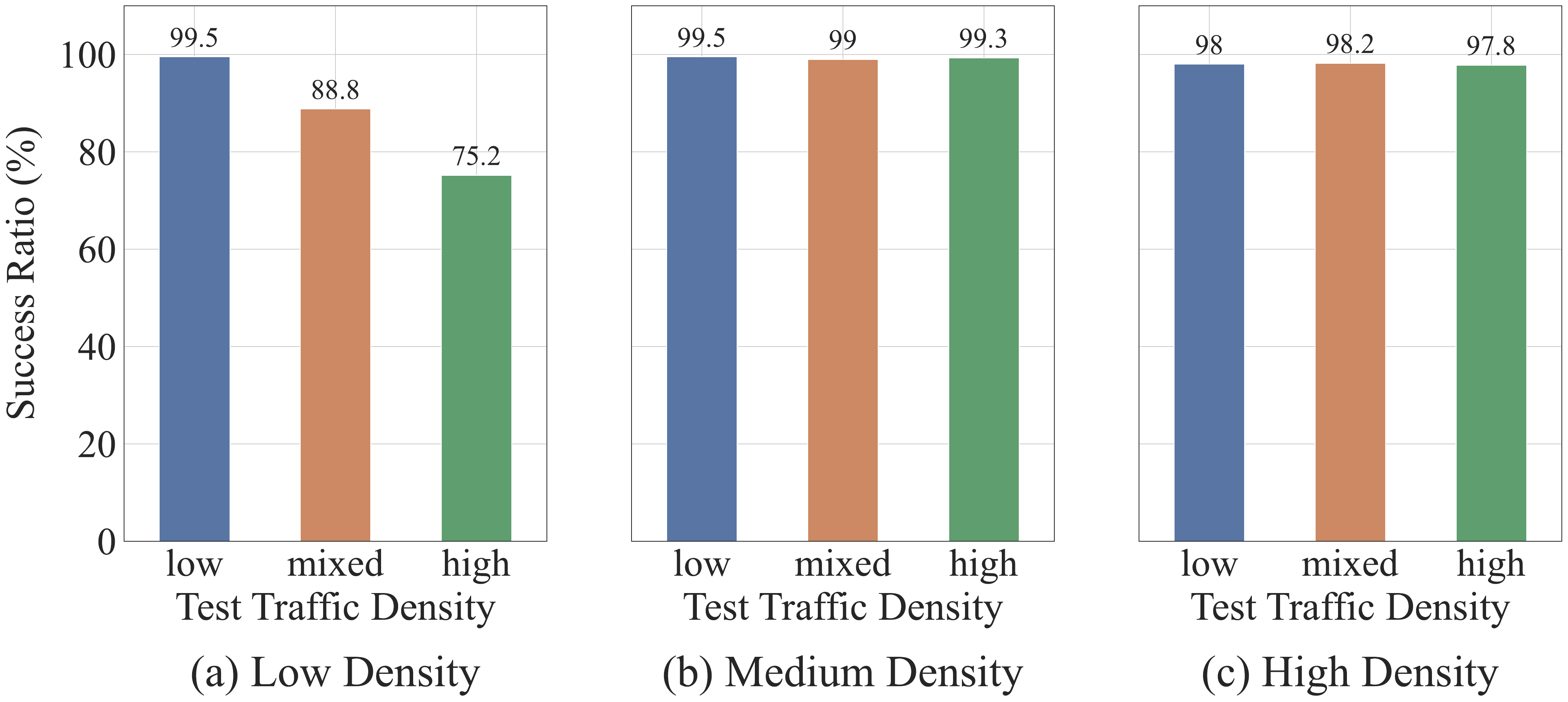}
    \caption{The success ratio of the trained agent in different densities of traffic. (a) low-density trained agents, (b) medium-density trained agents, and (c) high-density trained agents.}
    \label{fig:generalization ability}
\end{figure}

We evaluate the generalization ability of the proposed method by testing the agents in different densities of traffic. Fig. \ref{fig:generalization ability} (a) shows the success ratio of agents trained in the low density and we evaluate it in the low, medium, and high traffic densities, respectively. The highest success ratio occurs in the low traffic, which is 99.5$\%$. Fig. \ref{fig:generalization ability} (b)) and (c) show the medium-density and high-density trained agents, which have similar success ratios when tested in different densities of traffic, indicating a good generalization ability. In particular, the agent trained in the medium traffic density has the best performance and thus we train the agents in the medium traffic for evaluations. 
%%  figure: generalization ability

\subsection{Discussion}
{This study aims to investigate the combination of RL and MPC for the decision-making and planning of the autonomous on-ramp merging task. We propose a human-aligned safe RL approach with the consideration of personal human risk perception, which can adjust the safety level of the learned policy according to users' risk preferences. A Lagrangian-based SAC algorithm is employed to solve the high-level CMDP problem, followed by an MPC-based approach for low-level motion planning. To mask out unsafe RL actions, an ASM is built by pre-executing the RL action with MPC and conducting collision checks to determine whether the action is safe or not. Theoretical proof of the action shielding mechanism has shown its effectiveness in enhancing the safety and sampling efficiency of the RL framework. Simulation experiments in multiple densities of traffic have demonstrated that our method can enhance safety without sacrificing traffic efficiency and also reduce the computation burden. Different from existing studies, our method is built upon safe RL principles with a constraint setting related to user risk preference, which can reduce safety violations during the training process as well. The difficulty in balancing safety and other objectives can also be avoided due to the use of CMDP. Besides, the learned RL policy can adjust the safety level to meet the user's requirements and thus help improve public trust in AVs.
Besides, we conduct a convergence analysis to provide theoretical proof of the effectiveness of the action shielding, which is often ignored in most studies.}

{However, there are some limitations in the proposed method. For instance, the kinematic model used in our method is simple and ignores the effects of forces, friction, or inertia, which is not suitable for precise vehicle control in highly dynamic road conditions. Instead, the dynamic model describes how a system moves with applied forces and torques, which could be used in our future research on complex and dynamic scenarios, such as high-speed autonomous driving. Besides, future works aim to validate our method using large amounts of real-world driving data to evaluate the reliability and safety level of our approach. The environmental uncertainties could be considered by formulating a partially observable Markov decision process (POMDP) for the RL framework. We are also interested in improving the generalization ability of our method and the popular meta-learning framework, such as Model-Agnostic Meta-Learning (MAML) \cite{finn2017model}, which could be used to train agents on a variety of tasks, such that it can quickly adapt to new tasks or environments with few gradient steps and training samples. Moreover, the complex interactions with surrounding vehicles can be modeled using graph neural networks to help make proactive decisions based on the trajectory predictions of all related traffic participants. }

\section{Conclusion}
\label{conclusion}
This study proposes a human-aligned safe RL framework for the decision-making of the on-ramp merging task of autonomous driving, in which RL is used for high-level decisions, followed by a low-level MPC. We formulate the high-level decision problem as a CMDP that represents safety in cost terms. Human risk preference is incorporated into the constraint settings of CMDP, which aims to adjust the safety level of RL policy based on the user's expectations. Besides, we use the fuzzy control method to compute the threshold of the cost limits based on human risk preferences and traffic density and a Lagrangian-based SAC is used to solve the CMDP. An action shielding mechanism is designed to remove unsafe RL actions, and we theoretically prove its effectiveness in enhancing safety and sample efficiency. Numerical simulations and theoretical analysis demonstrate the superiority of our method regarding success rate, collision rate, and average cost. As the safety violations can be reduced due to the CMDP and shielding, our method provides a promising solution for safe online learning of RL in real-world environments. Future works involve using a dynamic vehicle model for high-speed driving conditions and modeling the uncertainties of the environment into the safe RL frameworks with POMDP.

\bibliographystyle{IEEEtran}
\bibliography{IEEEabrv,reference}

\end{document}